\documentclass[letterpaper]{article} 
\usepackage[preprint]{aaai2027}  
\usepackage[hyphens]{url}  
\usepackage{graphicx} 
\usepackage{natbib}  
\usepackage{caption} 
\usepackage{algorithm}
\usepackage{algorithmic}

\usepackage{newfloat}
\usepackage{listings}
\DeclareCaptionStyle{ruled}{labelfont=normalfont,labelsep=colon,strut=off} 
\floatstyle{ruled}
\newfloat{listing}{tb}{lst}{}
\floatname{listing}{Listing}

\usepackage{booktabs}
\usepackage{amsmath}
\usepackage{multirow}
\usepackage{enumitem}
\usepackage{amssymb}
\title{Who Remains, What Changes: Identity Anchored Composed Gait Retrieval}
\author{
    Jingchen Fei\textsuperscript{\rm 1}, Zengbin Wang\textsuperscript{\rm 1}, Yukun Liu\textsuperscript{\rm 2}, Muyi Sun\textsuperscript{\rm 1}, Shibiao Xu\textsuperscript{\rm 1}, Man Zhang\textsuperscript{\rm 1}\corresponding
}
\affiliations{

    \textsuperscript{\rm 1} Beijing University of Posts and Telecommunications\\
    \textsuperscript{\rm 2} Huazhong University of Science and Technology\\
    \{feijingchen, wzb1, muyi.sun, shibiaoxu, zhangman\}@bupt.edu.cn\\
    c\_oconaliu@hust.edu.cn
}

\begin{document}

\maketitle

\begin{abstract}
Gait recognition has achieved remarkable progress, yet existing methods remain confined to rigid visual matching and often overlook the potential of natural language instructions for interactive retrieval. In this paper, we introduce \textbf{Composed Gait Retrieval (CoGR)}, a novel task that retrieves a target gait sequence based on a reference sequence and a natural language modification query. To address the absence of existing datasets for this task, we design an automated annotation pipeline powered by large vision-language models (VLMs) to construct the first gait-language datasets: Language-Augmented CCPG and Language-Augmented CASIA-B. Building on this,  we propose
\textbf{ComposeGait}, an identity-anchored composition framework designed to
prevent the identity drift that arises when generic composed retrieval follows
the instruction but returns the wrong person. Its Part-aware Identity
Adapter (PIA) aggregates multi-frame, part-aware identity evidence into a
sample-specific ID token. We inject the ID tokens into both branches of a shared Q-Former to 
preserve identity, while excluding the ID-token outputs from the final retrieval embeddings. Joint identity
and task-adapted composed-retrieval objectives optimize this space end to end. 
We evaluate ComposeGait on both benchmarks and show that it achieves the best R@1 among the compared methods, reaching 72.38\% on Language-Augmented CCPG and 83.61\% on Language-Augmented CASIA-B. These results establish ComposeGait as a strong baseline for CoGR. The datasets and code will be made publicly available.
\end{abstract}

\section{Introduction}

Gait recognition, a prominent biometric technology, aims to identify
individuals by their unique walking patterns from a distance. Driven by deep
learning, recent advances have achieved high accuracy in verifying identities
under constrained covariates such as carrying a bag or wearing a
coat~\cite{yuCasia-B,liIndepthExplorationPerson2023,fanOpenGaitComprehensiveBenchmark2025}.
However, real-world intelligent surveillance often involves dynamic and
interactive retrieval. Rather than simply asking \textit{``Who is this
person?''}, an investigator may issue a semantic instruction such as
\textit{``Find this person, but now wearing a blue shirt and trousers of a
different color.''} Traditional gait retrieval systems are constrained by a
rigid sequence-to-sequence paradigm: they penalize cross-covariate differences
and cannot understand natural-language modifications.

\begin{figure}[t]
    \centering
    \includegraphics[width=0.95\linewidth]{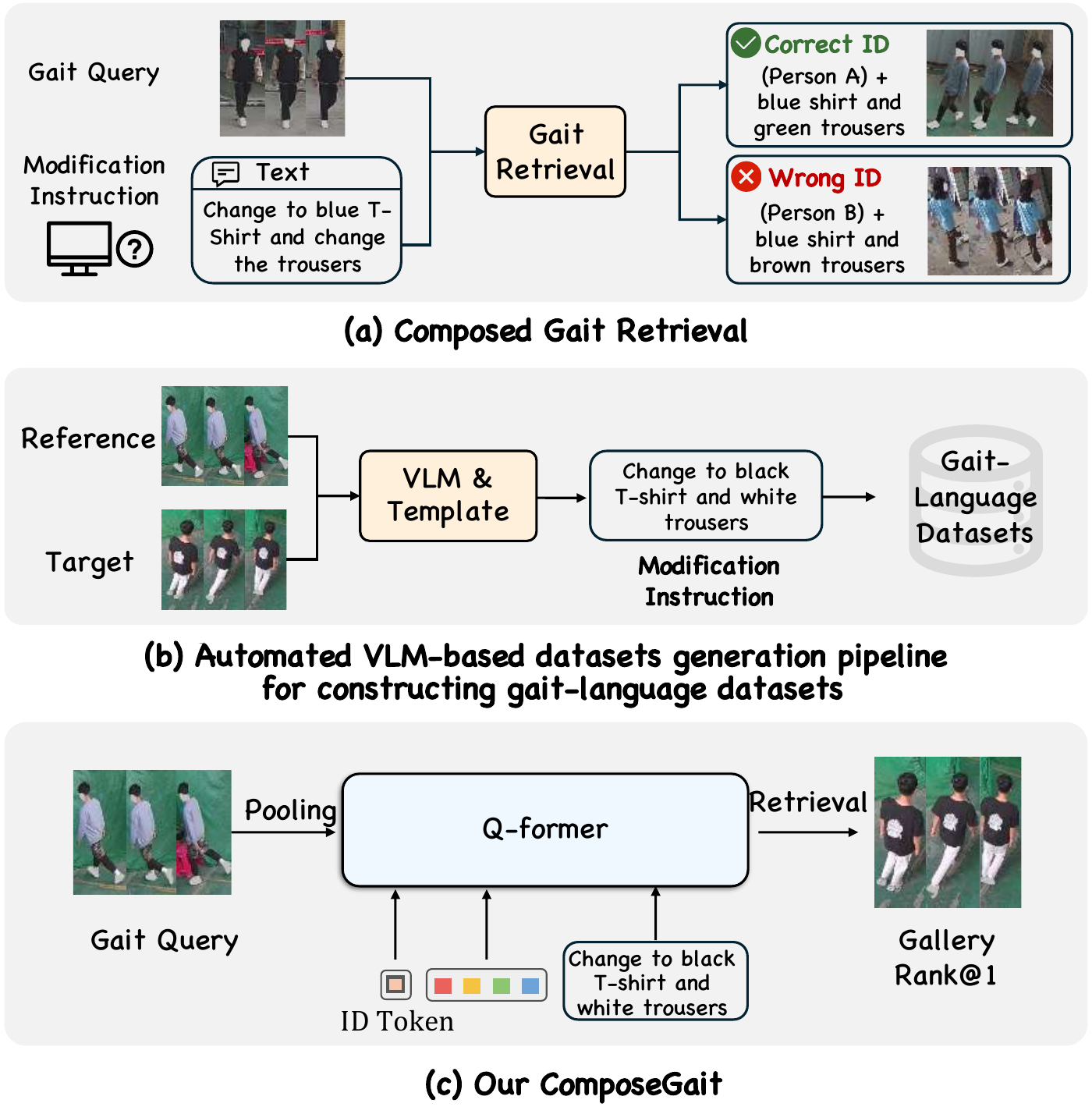}
    \caption{Overview of our contributions. (a) CoGR retrieves a target gait
    sequence that preserves identity while satisfying a relative instruction.
    (b) The automated VLM-based pipeline constructs gait-language training
    data. (c) Simplified illustration of our ComposeGait framework, which injects a sample-specific ID token into the Q-Former to anchor identity during semantic modification.}
    \label{fig:teaser}
\end{figure}

To bridge the modality gap between visual queries and user intentions,
Composed Image Retrieval (CIR)
\cite{duSurveyComposedImage2025,songComprehensiveSurveyComposed2025} has emerged
as a promising paradigm that retrieves target images from reference images and
relative text instructions. While successful in general image domains,
directly adapting existing CIR frameworks to gait encounters a critical
bottleneck: \textbf{identity--semantic entanglement}. Unlike static images, a
gait sequence is a high-dimensional spatiotemporal representation in which biometric
identity and appearance attributes are deeply intertwined. Mainstream CIR
methods typically fuse text and visual features early in the network to predict
a combined target embedding. In the gait context, such implicit composition
offers no explicit safeguard for the biometric cues that must remain stable.
As illustrated in Figure~\ref{fig:teaser}(a), this can lead to \textbf{identity
drift}: the model successfully modifies target attributes (e.g., retrieving a subject wearing a coat) but fatally fails to preserve the reference identity (e.g., retrieving the wrong subject). This failure mode is distinct from standard CIR,
where biometric identity preservation is not a requirement, and motivates an
identity-aware composition mechanism rather than generic feature fusion alone.

\begin{figure*}[t]
    \centering
    \includegraphics[width=0.9\textwidth]{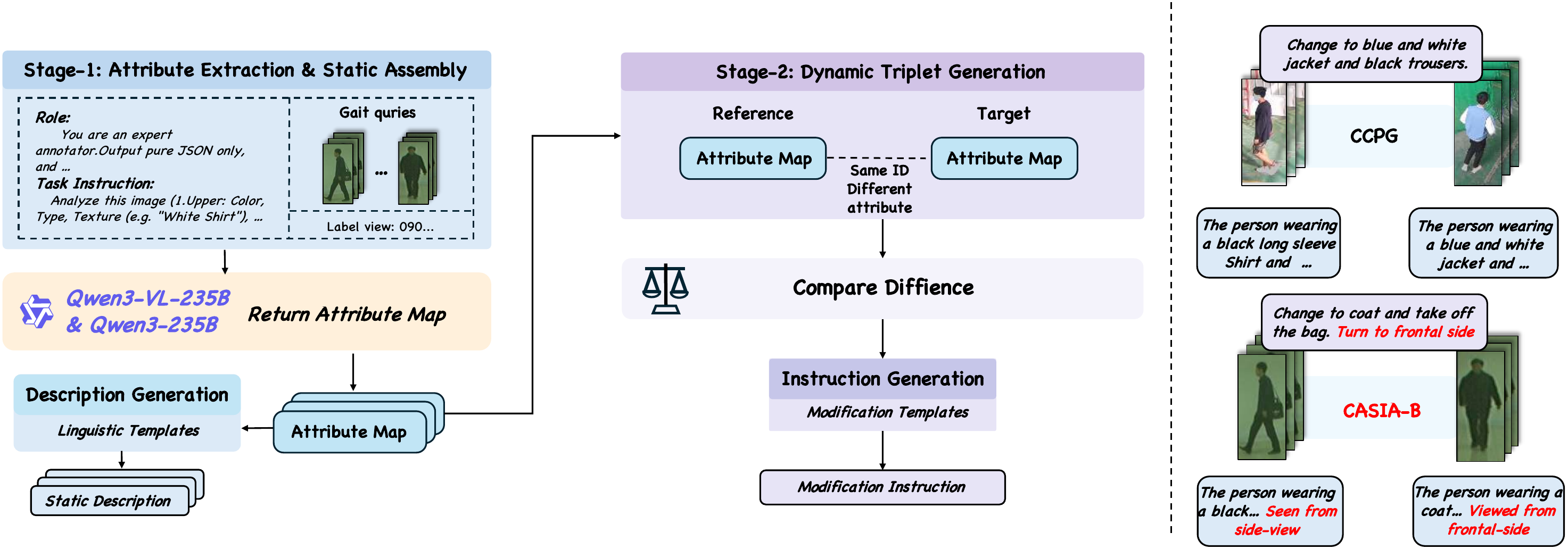}
    \caption{The automated VLM-based dataset construction pipeline. The first
    two stages extract fine-grained visual attributes and assemble static
    descriptions. Reference and target tracklets of the same identity are then
    compared to identify changed attributes and synthesize relative
    instructions. Language-Augmented CASIA-B includes explicit viewpoint shifts, whereas Language-Augmente CCPG
    focuses on appearance changes to reflect the structure of each dataset.}
    \label{fig:pipeline}
\end{figure*}

To overcome these challenges, we formally propose a novel and practical task,
namely \textbf{Composed Gait Retrieval (CoGR)}. Given a reference gait sequence
and a natural-language modification instruction, CoGR retrieves gallery
sequences that simultaneously preserve the reference identity and satisfy the
specified attribute changes. A key obstacle to this task is the absence of
gait-language training data: existing gait datasets provide only discrete
alphanumeric condition labels rather than natural-language descriptions. To
address this, we develop an automated annotation pipeline powered by large
Vision-Language Models (VLMs). As depicted in
Figure~\ref{fig:pipeline}, the pipeline dynamically translates discrete
experimental conditions into structurally consistent modification
instructions, establishing the \textbf{Language-Augmented CCPG} and
\textbf{Language-Augmented CASIA-B} datasets to benchmark this new paradigm.

With these datasets enabling CoGR training, we propose \textbf{ComposeGait}.As illustrated 
in Figure~\ref{fig:composegait}(a),ComposeGait is an identity-anchored composition 
framework built on BLIP-2. A frozen ViT-G
encodes sampled gait frames, while a \textbf{Part-aware Identity Adapter
(PIA)}(see Figure~\ref{fig:composegait}(b)) extracts a part-aware identity representation and projects it into the
Q-Former hidden space as a sample-specific \textbf{Identity token (ID token)}.
Appended to the pretrained queries, this token conditions multimodal reasoning
on gait identity.

ComposeGait injects ID tokens into both retrieval sides: the composed-query
branch processes the reference sequence and modification text with the reference
ID token, while the target branch processes each candidate with its ID token in
a shared Q-Former. Only the original query-token outputs form the retrieval
embeddings, allowing the ID tokens to guide attention without directly entering
the embedding. Joint learning combines identity classification and hard-triplet
losses for PIA with a task-adapted CoGR contrastive loss. The latter supports
multiple relevance-consistent positives and excludes same-identity targets that
violate the requested condition from the denominator, reducing false-negative
and identity-conflicting gradients.The main contributions of our work are summarized as follows:
\begin{itemize}
    \item We introduce Composed Gait Retrieval (CoGR), a novel
    interactive retrieval task. To support it, we develop an automated
    VLM-based pipeline to construct the first benchmarks for CoGR: Language-Augmented CCPG and
    Language-Augmented CASIA-B.
    \item We propose ComposeGait, an identity-anchored composition framework integrating a Part-aware Identity Adapter (PIA). This module projects part-aware gait features into sample-specific ID tokens and bidirectionally embeds them into a shared Q-Former. We also tailor the contrastive loss for CoGR matching, enabling multiple valid positives and filtering out same-identity targets with mismatched conditions in the loss denominator.
    \item Extensive experiments on the two language-augmented benchmarks
    evaluate retrieval effectiveness, identity preservation, and instruction
    satisfaction, establishing ComposeGait as a strong baseline for CoGR.
\end{itemize}

\section{Related Work}

\subsection{Gait Recognition}

Gait recognition has evolved from hand-crafted feature methods to deep
learning-based approaches~\cite{shenComprehensiveSurveyDeep2025}. Early
set-based methods such as GaitSet~\cite{chaoGaitSetRegardingGait2018} treat a
gait sequence as an unordered frame set, providing permutation invariance while
discarding explicit temporal order. The OpenGait framework
\cite{fanOpenGaitRevisitingGait2023,fanOpenGaitComprehensiveBenchmark2025}
systematically revisited existing designs and developed the structurally simple
yet effective GaitBase baseline, showing that practical recognition benefits
from both spatial and temporal modeling. Deeper architectures
\cite{fanExploringDeepModels2024} and dynamic aggregation mechanisms
\cite{maDANetDynamicAggregated2023} further capture discriminative motion
patterns in challenging outdoor scenarios.

Silhouettes remain the mainstream input modality, while alternative representations supply supplementary visual cues. RGB-based methods retain abundant appearance details yet suffer from sensitivity to clothing changes and potential privacy risks. GaitParsing~\cite{wangGaitParsingHumanSemantic2024} and ParsingGait~\cite{zhengParsingAllYou2023} leverage human semantic parsing to explicitly model separate body regions. GaitRef~\cite{zhuGaitRefGaitRecognition2023} optimizes skeleton sequences via silhouette temporal consistency, and LidarGait \cite{shenLidarGaitBenchmarking3D2023,shen2025lidargait++} leverages 3D point-cloud geometric features to achieve reliable outdoor gait recognition. Moving past traditional set-level and sequence-level modeling paradigms, GaitSnippet \cite{houGaitSnippetGaitRecognition2025} formulates gait as personalized action fragments, and SkeletonGait~\cite{fanSkeletonGaitGaitRecognition} transforms raw pose coordinates into silhouette-resembling skeleton maps. BigGait and BiggerGait \cite{yeBiggaitLearningGait2024,yeBiggerGaitUnlockingGait2025} further demonstrate that large vision models yield transferable gait embeddings, which motivates our PIA module to convert ViT-G features into part-aware identity representations.

The evolution of gait recognition has also been driven by increasingly
challenging datasets. CASIA-B~\cite{yuCasia-B} and
OU-MVLP~\cite{takemuraMultiViewLargePopulation2018} provide controlled indoor
settings with multiple viewpoints and limited covariates. Gait3D
\cite{zhengGaitRecognitionWild2022} and GREW~\cite{zhu2021gait} introduce
outdoor variations in viewpoint, occlusion, and illumination. Cross-covariate
benchmarks such as CCPG~\cite{liIndepthExplorationPerson2023} and
CCGR~\cite{zouCrossCovariateGaitRecognition2024} emphasize clothing and carrying
changes, while SUSTech1K~\cite{shenLidarGaitBenchmarking3D2023} provides
synchronized camera and LiDAR data. Despite this progress, existing methods and
benchmarks share the same objective: they suppress clothing, carrying, and
viewpoint variation as nuisance factors to recover an identity label. They
therefore retain a rigid 1:1 identity-matching formulation and cannot express
natural-language changes. CoGR changes the role of these covariates: identity
must remain invariant, but a user-specified subset of conditions becomes part
of the desired target. This extends gait analysis from passive identification
to identity-preserving, language-guided compositional retrieval.

\subsection{Composed Image and Video Retrieval}
Composed Image Retrieval (CIR) enables users to retrieve target images using a
multimodal query comprising a reference image and a modification text
\cite{duSurveyComposedImage2025,songComprehensiveSurveyComposed2025}.

\noindent\textbf{Image-Centric Composition.}
Supervised CIR learns joint representations from annotated triplets. TIRG
\cite{voComposingTextImage2019} uses residual gating, the CLIP-based Combiner
\cite{baldratiConditionedComposedImage2022} fuses reference and text features
with contrastive learning, and TG-CIR~\cite{wenTargetGuidedComposedImage2023}
and ConText-CIR~\cite{xingConTextCIRLearningConcepts2025} strengthen
target-guided and concept-level alignment. SPRC~\cite{bai2024sentence} instead
uses a BLIP-2 Q-Former to convert the reference--instruction pair into a
sentence-level prompt for text-based retrieval. Zero-shot methods reduce or
remove triplet supervision: SEARLE~\cite{baldratiZeroShotComposedImage2023} and
Pic2Word~\cite{saitoPic2WordMappingPictures2023} map the reference into CLIP's
textual space, while FreeDom~\cite{efthymiadisComposedImageRetrieval},
HyCIR~\cite{jiangHyCIR2024}, and BASIC~\cite{psomasInstanceLevelCIR2025} explores training-free fusion, while its i-CIR benchmark considers instance-level retrieval, which is related to identity preservation in CoGR. Despite
their different supervision regimes, these methods compose semantics in static
images without enforcing sequence-level biometric identity preservation.

\noindent\textbf{From Images to Video.}
CoVR-2~\cite{venturaCoVR2AutomaticData2024} automatically constructs video
triplets from caption differences through an LLM-assisted pipeline, whereas
FDCA~\cite{wuLEARNINGFINEGRAINEDREPRESENTATIONS2025} explores textual-token
disentanglement for fine-grained composed video retrieval. Although they extend
composition across frames, their retrieval objectives focus on object- or
scene-level edits rather than preserving biometric identity throughout a gait
sequence.

\noindent\textbf{Identity Beyond Static Appearance.}
Composed Person Retrieval (CPR) combines visual and textual queries to identify
people in image galleries. FAFA~\cite{liuAutomaticSyntheticData2025b} develops
LLM-assisted data synthesis and Q-Former-based fine-grained alignment. Its data
strategy is relevant to our automated construction pipeline, but CPR operates
on static person images. CoGR instead retrieves gait tracklets, for which
identity evidence is distributed across frames and must remain stable while
the requested appearance or viewpoint changes are applied. Static-image CPR
does not aggregate this multi-frame gait evidence or condition both query and
target encoding on sample-specific identity signals through a shared semantic
encoder.

Taken together, CIR supplies instruction-guided composition, composed video
retrieval extends it across frames, and CPR introduces person-level search.
CoGR is needed at their intersection: a valid target must follow the
instruction, preserve biometric identity over a gait sequence, and retain every
unspecified condition.

\begin{figure*}[t]
    \centering
    \includegraphics[width=1\textwidth,pagebox=cropbox]{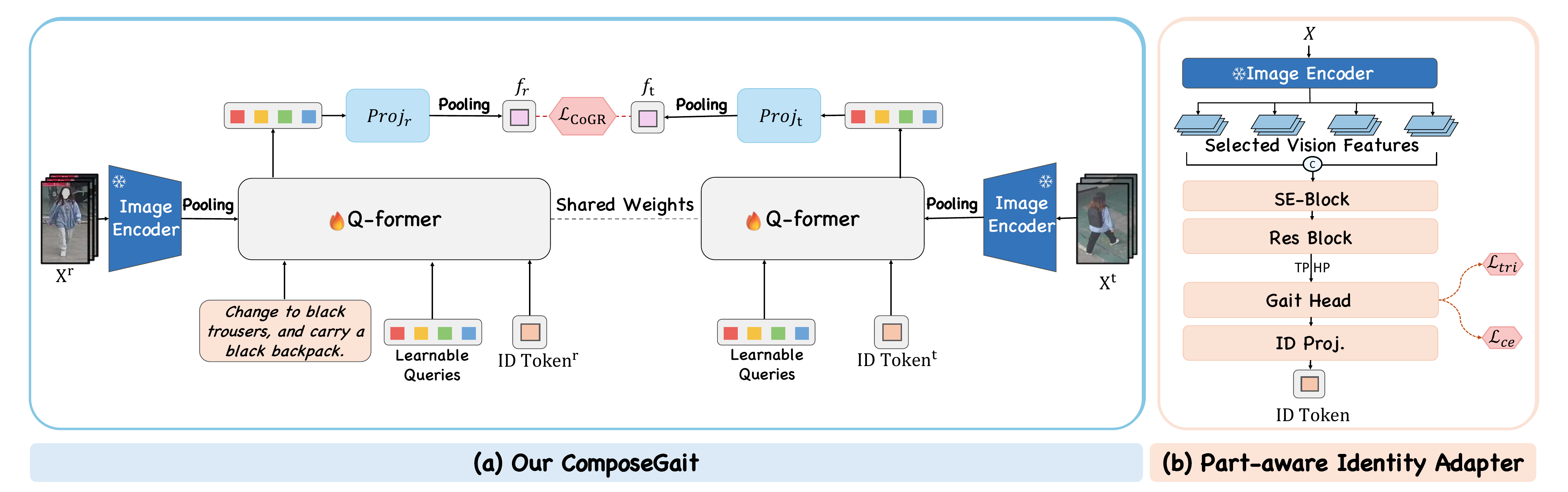}
    \caption{Overview of our ComposeGait and Part-aware Identity Adapter (PIA).
    (a) We adopt two Q-Formers with shared weights to process the reference gait
    sequence with its ID token and modification text, and the target gait
    sequence with its ID token. We optimize the resulting features using
    $\mathcal{L}_{\mathrm{CoGR}}$ during the training stage and match them for
    retrieval during inference. (b) We design PIA to extract part-aware identity
    representations for constructing the ID tokens.}
    \label{fig:composegait}
\end{figure*}
\section{Methodology}

\subsection{Paradigm Definition}
\label{sec:paradigm}

CoGR inherits the standard input--output form of composed retrieval
\cite{duSurveyComposedImage2025,songComprehensiveSurveyComposed2025}. Given a
reference gait sequence $X^r$, a natural-language modification $m$, and a
gallery $\mathcal{G}$, the model encodes the reference--text composition into a
query embedding $f_r$ and each candidate sequence $X\in\mathcal{G}$ into a
target embedding $f_t$. Both embeddings are
L2-normalized, and retrieval ranks candidates by their inner product:
\begin{equation}
    \hat{X}^{t}=\arg\max_{X\in\mathcal{G}}
    f_r(X^r,m)^{\top}f_t(X).
\end{equation}
Here, $\hat{X}^{t}$ denotes the top-ranked target sequence.
The gait-specific distinction lies in the relevance criterion. Let $y(X)$
denote subject identity and $c(X)$ the non-identity condition state (e.g., clothing, carrying status, viewpoint). The modification operator
$T_m$ updates only the components specified by $m$ and keeps every unspecified
component unchanged. The desired condition $c^t$ and the relevant target set are

\begin{equation}
\label{eq:cogr_relevance}
\begin{aligned}
    c^t &= T_m\!\left(c(X^r)\right),\\
    \mathcal{P}(X^r,m)
    &=\left\{X\in\mathcal{G}\ \middle|\
    \substack{y(X)=y(X^r)\\c(X)=c^t}\right\}.
\end{aligned}
\end{equation}
Accordingly, each training triplet $(X^r,m,X^t)$ uses a target
$X^t\in\mathcal{P}(X^r,m)$. The first condition enforces identity preservation,
whereas the second enforces both the requested change and preservation of all
unspecified conditions. CoGR therefore retains the composed-retrieval
formulation while making biometric identity part of the target definition.

\subsection{Automated Gait-Language Alignment Pipeline}
\label{sec:pipeline}

Existing gait datasets such as CASIA-B~\cite{yuCasia-B} and
CCPG~\cite{liIndepthExplorationPerson2023} were designed primarily for unimodal
identity verification, with carrying status, clothing, and camera conditions
encoded as discrete alphanumeric labels. Such labels lack the syntactic
structure and semantic richness required for natural-language-guided retrieval.
We bridge this gap with an automated annotation pipeline that uses a large VLM
to convert discrete condition states into fine-grained textual descriptions.

The pipeline operates in three sequential stages: attribute extraction, static
assembly, and dynamic triplet generation. We first query
Qwen3-VL-235B~\cite{qwen_vl} with a structured prompt containing semantic slots
such as
\texttt{\{upper\}}, \texttt{\{lower\}}, and \texttt{\{bag\}}. The VLM analyzes
sampled visual tracklets and fills these slots with fine-grained attributes.
During static assembly, the parsed attributes populate linguistic templates to
form an appearance description. For CASIA-B, an additional prompt converts
numerical camera angles into human-readable viewpoint expressions such as ``a
profile view,'' whereas CCPG uses only appearance attributes at this stage.

To construct composed-retrieval triplets, we pair reference and target tracklets
from the same identity and compare their attributes to identify clothing,
accessory, and viewpoint changes. The detected differences populate
modification templates to produce relative instructions. Both datasets share
the appearance-change templates, while CASIA-B adds slots for viewpoint shifts.
Because CASIA-B provides controlled camera angles but CCPG's arbitrary
surveillance-camera identifiers have no user-facing semantic meaning, we omit
viewpoint prompts for CCPG. This dataset-specific, rule-guided VLM-assisted
design keeps the queries visually grounded and structurally consistent at
scale.

\subsection{ComposeGait Overview}
\label{sec:overview}

ComposeGait implements identity-anchored composition with four coupled steps.
A frozen ViT-G encodes the sampled RGB frames once. PIA extracts a part-aware
identity representation from the sequence. A projector maps that
representation to one ID token. The token serves as the identity anchor and is
appended after the pretrained BLIP-2 query tokens. Finally, a shared Q-Former
encodes both the composed query and the
target. Only the outputs corresponding to the original pretrained queries are
projected, mean-pooled over the $M$ query tokens, and normalized to form the
retrieval embeddings $f_r$ and $f_t$.
Figure~\ref{fig:composegait}(a) summarizes this data flow.

\subsubsection{Part-aware Identity Adapter and ID Token Construction}
\label{sec:pia}
\label{sec:id_token}

As shown in Figure~\ref{fig:composegait}(b), PIA and the identity projector form a
continuous pathway from sequence-level identity evidence to a sample-specific
ID token. Before token construction, PIA builds on the part-aware identity
backbone of GaitBase~\cite{fanOpenGaitRevisitingGait2023} and the layer-wise
feature aggregation of BiggerGait~\cite{yeBiggerGaitUnlockingGait2025}. It
extracts hierarchical
features from selected layers of the frozen ViT-G~\cite{li2023blip}, applies
multi-layer fusion with an SE block and a residual
block~\cite{hu2018squeeze,he2016deep}, and then performs temporal max pooling
and horizontal pooling over $P$ body regions. A dedicated gait head maps the
pooled features to
$z^{\mathrm{id}}\in\mathbb{R}^{D_i}$, where $D_i$ denotes the identity-feature
dimension. The same pathway produces $z_r^{\mathrm{id}}$ and
$z_t^{\mathrm{id}}$ for the reference and target sequences, respectively.


ComposeGait then repurposes each identity representation as a conditioning token
rather than a standalone retrieval descriptor. A learned projector and token-type
embedding map it to the Q-Former hidden dimension:
\begin{equation}
    a=W_{\mathrm{id}}z^{\mathrm{id}}+b_{\mathrm{id}}+e_{\mathrm{type}},
    \qquad a\in\mathbb{R}^{D_q}.
\end{equation}
Here, $W_{\mathrm{id}}$ and $b_{\mathrm{id}}$ parameterize the identity
projector, $e_{\mathrm{type}}$ is the ID-token type embedding, and $D_q$ is the
Q-Former hidden dimension. This mapping produces the reference and target ID
tokens $a_r$ and $a_t$, each of which is appended after the $M$ pretrained
learnable queries $Q\in\mathbb{R}^{M\times D_q}$. The token-type embedding
distinguishes the identity anchor from the semantic queries, while the token
content remains sample-specific. This single-token bottleneck lets identity
influence attention without directly concatenating the full part representation
with the retrieval output.

\subsubsection{Bilateral Shared Q-Former}
\label{sec:bilateral}

The two retrieval branches use the same Q-Former weights but receive different
inputs. The composed-query branch processes the pretrained query tokens $Q$ and
reference ID token $a_r$ together with the reference visual tokens and
modification text. The target branch processes $Q$ and the target ID token $a_t$
together with the target visual tokens, without text. Weight sharing places the
two branches in the same embedding space, while $a_r$ and $a_t$ condition their
attention on the identity of the corresponding gait sequence.

The outputs at the $M$ original query-token positions are projected, mean-pooled,
and L2-normalized to form the query and target embeddings $f_r$ and $f_t$. The
output at the additional ID-token position is excluded: the ID token guides
Q-Former attention but is not directly included in the final retrieval embedding,
as illustrated in Figure~\ref{fig:composegait}(a).

\subsection{Joint Identity and Composition Learning}
\label{sec:objective}

PIA is supervised in the same batchwise manner as gait recognition. For a
mini-batch of $B$ reference--target pairs, let $Z_r^{\mathrm{id}}$ and
$Z_t^{\mathrm{id}}$ collect the corresponding identity representations, and
let $Y_r$ and $Y_t$ be their identity labels. We concatenate the two sides into
one gait-recognition batch,
\begin{equation}
    Z^{\mathrm{id}}=[Z_r^{\mathrm{id}};Z_t^{\mathrm{id}}],
    \qquad Y=[Y_r;Y_t],
\end{equation}
and compute identity classification and hard-triplet losses jointly:
\begin{equation}
\begin{aligned}
    \mathcal{L}_{\mathrm{id}}=
    {}&\mathcal{L}_{\mathrm{ce}}\!\left(h_{\mathrm{id}}(Z^{\mathrm{id}}),Y\right)+\mathcal{L}_{\mathrm{tri}}\!\left(Z^{\mathrm{id}},Y\right),
\end{aligned}
\end{equation}
where $h_{\mathrm{id}}$ is the identity classifier. Reference and target
samples therefore share the same classification batch and the same
positive/negative mining pool.
Standard pairwise contrastive learning treats only the target originally paired
with each query as positive. CoGR relevance is not necessarily one-to-one:
multiple in-batch targets may satisfy the same relevance criterion, and treating
such unpaired targets as negatives would introduce false-negative gradients.
Conversely, a target that shares the desired identity but violates a required
condition is invalid for the composed query yet remains positive in the identity
dimension. Treating it as an ordinary negative can therefore conflict with
identity preservation.

We address this asymmetric supervision by constructing the positive and
denominator sets from the CoGR relevance criterion. Let $\mathcal{P}(i)$ contain
the in-batch targets that satisfy Equation~\eqref{eq:cogr_relevance} for query
$i$. Here, $y_i$ is the reference identity associated with query $i$, and $y_j$
is the identity of target $j$. Same-identity targets outside $\mathcal{P}(i)$
form the ambiguous set $\mathcal{A}(i)$ and are excluded rather than treated as
different-identity negatives. Within a globally gathered batch of $B$
query--target pairs, we define
\begin{equation}
\begin{aligned}
\mathcal{A}(i)&=\{j\mid y_j=y_i,\ j\notin\mathcal{P}(i)\},\\
\mathcal{D}(i)&=\{1,\ldots,B\}\setminus\mathcal{A}(i).
\end{aligned}
\end{equation}
Here, $\mathcal{A}(i)$ is excluded from the denominator, whereas targets with
different identities remain in $\mathcal{D}(i)$ as negatives. Thus, the
positive supervision follows CoGR relevance while the denominator avoids
identity-conflicting negatives. For query $i$, the CoGR loss is
\begin{equation}
\begin{aligned}
\mathcal{L}_{\mathrm{CoGR}}^{(i)}={}&
-\frac{1}{|\mathcal{P}(i)|}
\sum_{p\in\mathcal{P}(i)}\log\frac{\exp(f_{r,i}^{\top}f_{t,p}/\tau)}
{\sum_{j\in\mathcal{D}(i)}
\exp(f_{r,i}^{\top}f_{t,j}/\tau)}.
\end{aligned}
\end{equation}
Here, $f_{r,i}$ denotes the embedding of query $i$, $f_{t,j}$ denotes the
embedding of target $j$, $p$ indexes a positive target in
$\mathcal{P}(i)$, and $\tau$ is the temperature. All embeddings are
L2-normalized. During training, $\mathcal{L}_{\mathrm{CoGR}}$ denotes the
average of $\mathcal{L}_{\mathrm{CoGR}}^{(i)}$ over the queries in the batch.
The complete objective is
\begin{equation}
    \mathcal{L}=\mathcal{L}_{\mathrm{CoGR}}+
    \lambda_{\mathrm{id}}\mathcal{L}_{\mathrm{id}}.
\end{equation}

\section{Experiments}

\subsection{Experimental Setup}
\begin{table*}[t]
    \centering
    \small
    \setlength{\tabcolsep}{1.1mm}
    \begin{tabular}{@{}llcccccccccc@{}}
    \toprule
    \multirow{2}{*}{Method} & \multirow{2}{*}{Backbone} &
    \multirow{2}{*}{Task} & \multicolumn{4}{c}{CCPG} &
    \multicolumn{5}{c}{CASIA-B} \\
    \cmidrule(lr){4-7}\cmidrule(lr){8-12}
    & & & R@1 & R@5 & R@10 & ID R@1 & R@1 & SC-R$_a$@1 &
    SC-R$_v$@1 & SC-R$_c$@1 & ID R@1 \\
    \midrule
    Text Only & ViT-B/32 & CIR & 7.74 & 21.26 & 29.34 & 11.64 & 0.86 & 5.68 & 0.50 & 1.02 & 4.34 \\
    Image Only & ViT-B/32 & CIR & 3.96 & 10.91 & 15.62 & 26.74 & 2.13 & 10.25 & 1.83 & 0.28 & 62.18 \\
    Image+Text & ViT-B/32 & CIR & 7.61 & 18.97 & 26.61 & 23.67 & 2.77 & 14.27 & 2.71 & 0.87 & 46.18 \\
    \midrule
    TIRG~\cite{voComposingTextImage2019} & RN18 & CIR & 14.71 & 32.41 & 42.14 & 29.21 & 7.73 & 33.66 & 11.61 & 5.13 & 58.80 \\
    TG-CIR~\cite{wenTargetGuidedComposedImage2023} & ViT-B/16 & CIR & 18.27 & 40.66 & 51.97 & 23.49 & 19.46 & 31.02 & 31.20 & 16.05 & 32.17 \\
    SEARLE~\cite{baldratiZeroShotComposedImage2023} & ViT-B/32 & CIR & 6.83 & 18.48 & 26.92 & 10.67 & 1.49 & 8.80 & 0.76 & 1.38 & 6.99 \\
    BASIC~\cite{psomasInstanceLevelCIR2025} & ViT-L/14 & CIR & 12.82 & 32.00 & 42.69 & 23.52 & 3.65 & 14.34 & 5.35 & 2.71 & 29.62 \\
    FreeDom~\cite{efthymiadisComposedImageRetrieval} & ViT-L/14 & CIR & 11.03 & 26.46 & 35.67 & 30.55 & 2.75 & 12.47 & 3.70 & 0.72 & 54.21 \\
    CLIP4CIR~\cite{baldratiConditionedComposedImage2022} & RN50 & CIR & 24.69 & 55.52 & 67.59 & 31.27 & 22.34 & 27.35 & 39.75 & 19.84 & 35.21 \\
    CoVR-2~\cite{venturaCoVR2AutomaticData2024} & ViT-G/14 & CoVR & 57.09 & 78.51 & 86.31 & 71.92 & 36.21 & \textbf{83.80} & 34.26 & 41.24 & \textbf{86.72} \\
    SPRC~\cite{bai2024sentence} & ViT-G/14 & CIR & 57.72 & 79.43 & 86.16 & 62.27 & \underline{66.46} & 71.40 & \underline{82.82} & \underline{61.16} & 71.57 \\
    FAFA~\cite{liuAutomaticSyntheticData2025b} & ViT-G/14 & CPR & \underline{68.82} & \textbf{89.79} & \textbf{93.73} & \underline{76.22} & 43.77 & 75.69 & 49.75 & 42.21 & 84.73 \\
    \midrule
    ComposeGait (Ours) & ViT-G/14 & CoGR & \textbf{72.38} & \underline{83.18} & \underline{87.44} & \textbf{76.56} & \textbf{83.61} & \underline{83.31} & \textbf{96.56} & \textbf{78.36} & \underline{84.84} \\
    \bottomrule
    \end{tabular}
    \vspace{0.5mm}
    \caption{Main comparison on Language-Augmented CCPG and CASIA-B (\%). Bold and
    underlined values indicate the best and second-best results in each column.}
    \label{tab:main_benchmark_comparison}
\end{table*}
\paragraph{Datasets.}
Language-Augmented CCPG~\cite{liIndepthExplorationPerson2023} contains 50,000
triplets, split into 40,000 for training and 10,000 for evaluation. It covers
diverse clothing and carrying changes in unconstrained scenes.
Language-Augmented CASIA-B~\cite{yuCasia-B} contains 64,506
triplets, with 55,160 for training and 9,346 for testing. It spans
$0^{\circ}$--$180^{\circ}$ viewpoints and normal, bag, and coat conditions
under a controlled multi-camera protocol. In both benchmarks, training and
evaluation identities are disjoint, and the original gallery definitions are
preserved.

\paragraph{Metrics.}
R@K counts a retrieval as correct only when the result preserves identity and
all unspecified covariates while satisfying the requested change. Specified-
Change Recall (SC-R@K) checks only whether the requested change is realized. On
CASIA-B, we additionally report attribute-only $\mathrm{SC\mbox{-}R}_a$,
viewpoint-only $\mathrm{SC\mbox{-}R}_v$, and composite
$\mathrm{SC\mbox{-}R}_c$. ID R@1 measures whether the top-ranked item has the
correct identity regardless of its condition. These metrics distinguish strict
task success from instruction satisfaction and identity preservation. All
rank-based metrics are reported in percent (\%); signed values denote absolute
gains.

\paragraph{Implementation details.}
ComposeGait initializes the vision encoder, Q-Former, query tokens, and
projection heads from BLIP-2~\cite{li2023blip}, using the
\texttt{Salesforce/blip2-itm-vit-g} checkpoint. Images are resized to
$224\times224$, and text is truncated to 64 tokens. For PIA, we use ViT-G
layers $\{8,16,24,39\}$ and set $P=16$ and $D_i=768$. We sample at most 30 frames per sequence during training and at most 60 frames during evaluation. The vision encoder and text embeddings are frozen. We train the
Q-Former, pretrained query tokens, query/target projection heads, PIA, identity
projector, and ID-token type embedding. PIA uses triplet margin 0.3 and label
smoothing 0.1. We set the identity classifier with 90 classes on CCPG and 64 classes on CASIA-B, 
leaving 10 identities for validation. Training uses
AdamW with learning rate $2\times10^{-5}$, weight decay 0.05, 500 warmup steps,
and cosine decay for 20,000 iterations. Balanced batches contain 16 triplets
with four instances per identity. The contrastive temperature is 0.07,
$\lambda_{\mathrm{id}}=1$, and gradient norm is clipped at 1.0. All experiments run on a single NVIDIA RTX 5090 GPU.

\subsection{Comparison with Prior Methods}

We compare against naive baselines (Text Only, Image Only, and Image+Text),
supervised CIR models (TIRG~\cite{voComposingTextImage2019},
CLIP4CIR~\cite{baldratiConditionedComposedImage2022},
TG-CIR~\cite{wenTargetGuidedComposedImage2023}),and SPRC
\cite{bai2024sentence}, and zero-shot CIR methods
(SEARLE~\cite{baldratiZeroShotComposedImage2023},
BASIC~\cite{psomasInstanceLevelCIR2025}, and
FreeDom~\cite{efthymiadisComposedImageRetrieval}). We also include the
composed-video retrieval baseline CoVR-2
\cite{venturaCoVR2AutomaticData2024} and the composed-person retrieval method
FAFA~\cite{liuAutomaticSyntheticData2025b}. For every reported
model--dataset pair, training and tuning are conducted independently on the
corresponding language-augmented dataset. We report performance under the common
evaluation protocol, using the same identity-disjoint
splits and gallery definitions.
Table~\ref{tab:main_benchmark_comparison} reports the comparison on both
datasets.

\paragraph{Results on CCPG.}
ComposeGait achieves 72.38\% R@1, outperforming the same-backbone FAFA by 3.56
percentage points (pp), and obtains the best ID R@1 of 76.56\%. Although FAFA
performs better at R@5 and R@10, our R@1 gain shows stronger top-ranked matching
under the joint identity and instruction constraints.

\paragraph{Results on CASIA-B.}
Under severe cross-view variations, conventional supervised and zero-shot CIR
methods degrade substantially, reaching at most 22.34\% R@1. ComposeGait
achieves 83.61\% R@1, outperforming the strongest competitor SPRC by 17.15
pp. Notably, its 96.56\% viewpoint-only recall shows that ComposeGait
handles view changes particularly well, while the best composite recall of
78.36\% indicates that this advantage persists when viewpoint and attribute
changes must be satisfied simultaneously.

\subsection{Architectural Ablation Study}

We ablate Shared QF, PIA, and ViT$\rightarrow$QF Multi on CCPG. Shared QF
makes the query and target branches share one Q-Former; otherwise, they use
independently parameterized Q-Formers. PIA always aggregates multiple frames
in its identity pathway. ViT$\rightarrow$QF Multi controls the semantic
Q-Former path, aggregating valid-frame ViT tokens when enabled or using the
middle valid frame otherwise.
Table~\ref{tab:ccpg_architectural_ablation} reports retrieval performance and
parameter counts.

\begin{table}[t]
    \centering
    \setlength{\tabcolsep}{2.8pt}
    \footnotesize
    \begin{tabular}{@{}cccccc@{}}
    \toprule
    \shortstack{Shared QF} & PIA &
    \shortstack{ViT$\rightarrow$QF Multi} & R@1 & ID R@1 &
    \shortstack{Params.(B)} \\
    \midrule
    $\times$ & $\times$ & $\times$ & 56.71 & 60.08 & 1.33 \\
    \checkmark & $\times$ & $\times$ & 61.47 & 65.31 & 1.17 \\
    $\times$ & \checkmark & $\times$ & 57.52 & 61.77 & 1.36 \\
    \checkmark & \checkmark & $\times$ & 67.74 & 72.63 & 1.20 \\
    \checkmark & \checkmark & \checkmark & \textbf{72.38} & \textbf{76.56} & 1.20 \\
    \bottomrule
    \end{tabular}
    \caption{Architectural ablation experiment on Language-Augmented CCPG.}
    \label{tab:ccpg_architectural_ablation}
\end{table}

\paragraph{Importance of PIA-Based Identity Anchoring.}
With Q-Former sharing, PIA raises R@1 from 61.47\% to 67.74\% (+6.27 pp) and
ID R@1 from 65.31\% to 72.63\% (+7.32 pp). Without sharing, the gains shrink
to 0.81 pp and 1.69 pp. Because PIA adds the same 0.03 billion parameters in both cases,
this asymmetric benefit cannot be explained by increased capacity alone and
suggests that PIA is more effective when the query and target branches use the
same Q-Former weights.

\paragraph{Importance of Q-Former Sharing.}
Without PIA, sharing one Q-Former between the two branches raises R@1 from
56.71\% to 61.47\% (+4.76 pp) and ID R@1 from 60.08\% to 65.31\% (+5.23 pp).
With PIA, the gains reach 10.22 pp and 10.86 pp. With PIA, the gains
grow to 10.22 pp and 10.86 pp, respectively. In both settings, weight sharing
reduces the parameter count by 0.16 billion. These results favor a common
Q-Former parameterization across the query and target branches, particularly
when PIA is enabled, rather than attributing the gains to increased model size.

\paragraph{Importance of Multi-Frame Evidence.}
With Shared QF and PIA enabled, ViT$\rightarrow$QF Multi raises R@1 from
67.74\% to 72.38\% (+4.64 pp) and ID R@1 from 72.63\% to 76.56\% (+3.93 pp).
Both configurations use 1.20 billion parameters, showing that
the improvement is not due to increased model capacity and is consistent with
multi-frame aggregation providing stronger evidence to the semantic Q-Former.

\subsection{Controlled ID-Token Interventions}

We compare four ID-token configurations. Correct ID retains the original tokens
on both branches, whereas Zero ID and Random ID respectively zero and randomize
them. Shuffled Query ID replaces each query-side identity feature with one from
a different gallery identity while leaving the target-side ID tokens unchanged.
Table~\ref{tab:id_token_interventions} reports R@1 for strict CoGR success,
ID R@1 for top-ranked identity preservation, and mAP for overall ranking
quality; $\Delta$R@1 is measured relative to Correct ID.

\begin{table}[t]
\centering
\setlength{\tabcolsep}{2.8pt}
\footnotesize
\begin{tabular}{@{}lcccc@{}}
\toprule
Intervention & R@1 & $\Delta$R@1 & ID R@1 & mAP \\
\midrule
Correct ID & 72.38 & 0.00 & 76.56 & 69.64 \\
Zero ID & 69.44 & -2.94 & 73.30 & 65.84 \\
Random ID & 68.39 & -3.99 & 72.17 & 65.24 \\
Shuffled Query ID & 64.27 & -8.11 & 67.83 & 62.23 \\
\bottomrule
\end{tabular}
\caption{Controlled ID-token intervention on Language-Augmented CCPG.}
\label{tab:id_token_interventions}
\end{table}

\paragraph{Removing or Randomizing the ID Token.}
Zeroing both ID tokens reduces R@1 from 72.38\% to 69.44\% ($-2.94$ pp), ID R@1
from 76.56\% to 73.30\% ($-3.26$ pp), and mAP from 69.64\% to 65.84\%
($-3.80$ pp). Randomizing both tokens causes larger drops to 68.39\% R@1,
72.17\% ID R@1, and 65.24\% mAP. Compared with Zero ID, Random ID is lower by
1.05 pp, 1.13 pp, and 0.60 pp on the three metrics, respectively. This pattern
indicates that the model uses the content of the ID tokens: arbitrary identity
signals are more disruptive than removing the tokens altogether.

\paragraph{Shuffling Query-Side Identity.}
Shuffled Query ID produces the largest degradation, reaching 64.27\% R@1
($-8.11$ pp), 67.83\% ID R@1 ($-8.73$ pp), and 62.23\% mAP ($-7.41$ pp)
relative to Correct ID. Because the target-side tokens remain unchanged, this
result provides controlled evidence that the query-side anchor carries
sample-specific biometric information needed to align the composed query with
the correct gallery identity. These fixed-checkpoint interventions demonstrate
model dependence on the ID-token content.

\begin{figure}[t]
    \centering
    \includegraphics[width=\linewidth]{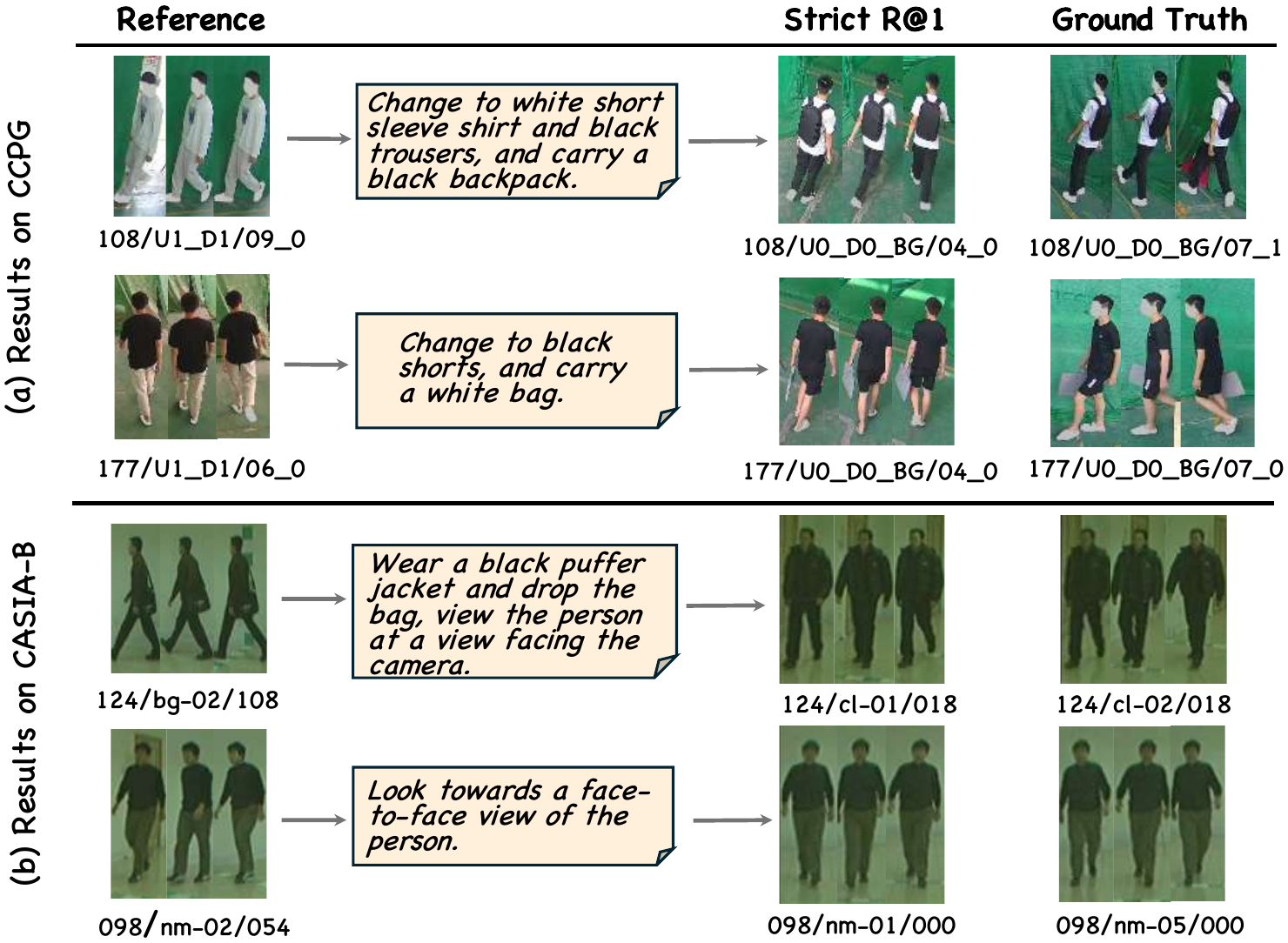}
    \caption{Qualitative results on (a) Language-Augmented CCPG dataset and (b) Language-Augmented  CASIA-B dataset.}
    \label{fig:Qualitative Results}
\end{figure}

\subsection{Qualitative Results}

Figure~\ref{fig:Qualitative Results} presents qualitative retrieval examples from Language-Augmented CCPG and CASIA-B. In panel (a), the retrieved CCPG sequences follow fine-grained appearance instructions (e.g., clothing and carrying status) while preserving reference identity. Panel (b) shows CASIA-B cases combining appearance modifications with viewpoint changes. Even during drastic perspective shifts (side to frontal), the retrieved sequences reflect both requested attributes. These examples qualitatively illustrate how ComposeGait successfully balances instruction compliance with identity preservation across various covariates.

\section{Conclusion}

We introduce Composed Gait Retrieval (CoGR) and an automated VLM-based
pipeline to construct Language-Augmented CCPG and Language-Augmented CASIA-B.
ComposeGait mitigates identity drift by generating sample-specific ID tokens
from part-aware, multi-frame gait features and injecting them into both
branches of a shared Q-Former. Extensive experiments demonstrate
state-of-the-art R@1 performance on both benchmarks.

Future work will extend CoGR toward large-scale in-the-wild datasets,
open-vocabulary descriptions, more flexible annotation strategies, and
identity-aware composition under unconstrained retrieval scenarios.
\bibliography{references}
\clearpage
\appendix

\section{VLM Prompts for Dataset Construction}
\label{sec:appendix_prompts}

To construct our composed gait-language datasets, we deployed an automated pipeline utilizing Large Vision-Language Models (VLMs), specifically Qwen3-VL-235B, to meticulously extract fine-grained visual attributes and synthesize diverse compositional instructions. The exact prompt templates designed for the distinct characteristics of the CASIA-B and CCPG datasets are provided below.

\subsection{CASIA-B Prompt Templates}
For the CASIA-B dataset, the extraction focuses on specific clothing elements and carried accessories, while the synthetic instructions explicitly articulate controlled viewpoint shifts.

\vspace{1mm}
\noindent\textbf{1. System Prompt for Modification Instruction Synthesis:}
\begin{lstlisting}
System Role: You are a Senior Computer Vision Data Architect and Linguistic 
Expert specializing in constructing robust datasets for Composed Image Retrieval.

Dataset Context: Subject: A pedestrian walking.

Placeholder Constraint (CRITICAL): Every single sentence MUST include the 
placeholder {subject} to represent the person (e.g., "{subject} walking").
Do not use "the person" or "a man/woman" directly; always use {subject}.

Requirements:
- Language: English, natural, native-speaker phrasing.
- Safety & Generalization: Strictly adhere to generic terms above to ensure 
  compatibility with mixed datasets (CASIA-B + CCPG).
- Linguistic Diversity: Since nouns are restricted, you must vary sentence 
  structures (imperative, descriptive, questioning).
- Quantity: Generate 100 unique sentences for each key.

Format: Output a single valid JSON object. No markdown, no intro text.

JSON Keys to Generate:
"change_view": Change the camera viewpoint. Constraint: MUST include both 
{subject} and {view} placeholders (e.g., "Switch to the {view} of {subject}.").
"connectors": A list of 15 conjunctions or transition phrases.
\end{lstlisting}

\vspace{1mm}
\noindent\textbf{2. Attribute Extraction Prompts:}
\begin{lstlisting}
SYSTEM_PROMPT = """You are an expert annotator. Output pure JSON only. 
NO conversational text."""

PROMPT_UPPER = """
Analyze this image. 
Task: Describe the upper clothing in 1 to 5 words ONLY (Color + Type).
Example outputs: "white t-shirt", "black long coat", "red jacket".
Output JSON:
{"upper": "your short description"}
"""

PROMPT_BAG = """
Analyze this image. 
Task: Describe the bag the person is carrying in 1 to 5 words ONLY (Color 
+ Type). Ignore the clothing.
Example outputs: "black backpack", "white handbag", "brown shoulder bag".
Output JSON:
{"bag": "your short description"}
"""
\end{lstlisting}
\vspace{1mm}
\noindent\textbf{3. View Description Prompts:}
\begin{lstlisting}
SYSTEM_PROMPT = """You are an expert annotator. Output pure JSON only. 
NO conversational text."""

PROMPT_VIEW = """
Analyze this image. 
Task: Describe the camera view angle relative to the pedestrian in 1 to 5 words ONLY. 
Constraint: DO NOT use full sentences or subjects (e.g., "The person is"). Use purely descriptive view phrases.
Example outputs based on view categories:
- Front: "frontal view", "captured head-on"
- Front-Side: "semi-frontal angle", "front-lateral view"
- Side: "full profile view", "pure side perspective"
- Back-Side: "semi-rear view", "oblique back angle"
- Back: "full rear view", "captured from behind"
Output JSON:
{"view": "your short description"}
"""
\end{lstlisting}

\subsection{CCPG Prompt Templates}
Collected from unconstrained surveillance environments, the CCPG dataset presents unique challenges. The prompts are strictly tailored to mask invisible regions (e.g., shoes) and handle occlusion gracefully.

\vspace{1mm}
\noindent\textbf{1. System Prompt:}
\begin{lstlisting}
SYSTEM_PROMPT = """
You are an expert data annotator for Surveillance Gait Recognition.
**RULES:**
1. **NO SHOES:** The feet/shoes area is MASKED/PAINTED OUT. Do NOT describe shoes.
2. **JSON ONLY:** Output pure JSON format. Do not use Markdown code blocks.
"""
\end{lstlisting}

\vspace{1mm}
\noindent\textbf{2. Appearance Extraction Prompt:}
\begin{lstlisting}
CLOTHING_PROMPT_TEMPLATE = """
Analyze this image (Best available view).
**Context:** Ground Truth says: {gt_bag_str}.

**Task:** Describe Appearance STRICTLY.
1. **Upper:** Color, Type, Texture (e.g. "Red Hoodie", "White Shirt").
2. **Lower:** Trousers/Shorts/Skirt ONLY (Ignore shoes).
3. **Bag:** Describe visual details (Color/Type) if visible.

**Output JSON:**
{
  "upper": "Visual description...",
  "lower": "Visual description...",
  "bag_visual": "Visual description..."
}
"""
\end{lstlisting}

\section{Manual Review of the Datasets}
\label{sec:Manual Review}

To ensure the high quality and semantic reliability of our synthesized composed gait-language datasets, we conducted a rigorous manual review process. Due to the vast scale of the generated datasets, it is computationally and temporally prohibitive to verify every single instance manually. Therefore, we adopted a robust random sampling strategy.

\textbf{Review unit.} 

We randomly sampled 500 reference--target--instruction triplets from each of the CASIA-B and CCPG datasets, resulting in 1,000 reviewed triplets in total. Each triplet consists of a source gait sequence, its corresponding static description, a modification instruction, a target gait sequence, and the corresponding target description. To reduce potential author bias, we recruited three independent human evaluators who were not involved in the development of the project.

\textbf{Evaluation criteria.}

The evaluators independently assessed whether the static descriptions accurately reflected the visual appearance of the corresponding gait sequences and whether the modification instructions accurately described the changes from the source to the target sequence. A triplet was considered fully correct only when all three evaluators agreed that all of its textual components were accurate.

\textbf{Review results.}

Table~\ref{tab:manual_review} summarizes the dataset statistics and manual review results. The results show that more than ($>92\%$) of the reviewed annotations satisfy the evaluation criteria, indicating that the automated VLM-based annotation pipeline produces generally reliable gait-language annotations. The remaining errors also highlight the need for appropriate quality control when using automatically generated annotations.

\begin{table}[h]
  \centering
  \begin{tabular}{lcc}
    \toprule
    \textbf{Metric} & \textbf{CASIA-B} & \textbf{CCPG} \\
    \midrule
    Generated Triplets& 64,506 & 50,000 \\
    Sampled Triplets for Review & 500 & 500 \\
    \textbf{Manual Verification Accuracy} & \textbf{92.6\%}& \textbf{96.0\%}\\
    \bottomrule
  \end{tabular}
  \caption{Manual Review Results for the Synthesized Datasets.}
  \label{tab:manual_review}
\end{table}

\section{Additional Ablation Studies}
\label{sec:appendix_additional_ablations}
\subsection{Bilateral ID-Token Injection}
\label{sec:appendix_bilateral_id}

We study whether the ID token should condition only one retrieval branch or
both branches. The composed-query-only variant appends the reference ID token
$a_r$ only to the query-side Q-Former, whereas the target-only variant appends
$a_t$ only to the target-side Q-Former. The bilateral variant injects the
corresponding token into both branches. We vary only the injection scope while
keeping the remaining architecture and training protocol unchanged.
Table~\ref{tab:appendix_id_injection_scope} reports the results.

\begin{table}[t]
    \centering
    \footnotesize
    \setlength{\tabcolsep}{4.5pt}
    \begin{tabular}{@{}lcc@{}}
        \toprule
        Injection scope & R@1 & ID R@1 \\
        \midrule
        Composed-query only ($a_r$) & 70.01 & 73.96 \\
        Target only ($a_t$) & 71.12 & 75.09 \\
        Bilateral ($a_r,a_t$) & \textbf{72.38} & \textbf{76.56} \\
        \bottomrule
    \end{tabular}
    \caption{Ablation of the ID-token injection scope on
    Language-Augmented CCPG (\%). Bilateral conditioning performs best on both
    strict retrieval and identity preservation.}
    \label{tab:appendix_id_injection_scope}
\end{table}

Conditioning either branch yields a viable identity-anchored representation,
but conditioning both sides is consistently stronger. Relative to
composed-query-only injection, bilateral injection improves R@1 by 2.37
percentage points (pp) and ID R@1 by 2.60 pp. It also exceeds target-only
injection by 1.26 pp and 1.47 pp, respectively. Target-only injection is itself
1.11 pp better in R@1 and 1.13 pp better in ID R@1 than composed-query-only
injection, suggesting that anchoring gallery representations is particularly
useful. The best performance nevertheless requires both $a_r$ and $a_t$:
conditioning the shared Q-Former on the identity evidence of each input places
the composed query and gallery candidates in a mutually identity-aware
embedding space, rather than leaving one side unanchored.

\subsection{Comparison of Contrastive Objectives}
\label{sec:appendix_loss_ablation}

We compare three contrastive objectives on Language-Augmented CASIA-B while
keeping the remaining setup unchanged. Their difference can be stated under a
common notation. For query $i$, let $p_i$ denote its designated ground-truth target,
$\mathcal{B}=\{1,\ldots,B\}$ the in-batch target indices, and
$\mathcal{P}(i)$ all targets satisfying the CoGR relevance criterion. We define
the same-identity but condition-mismatched set and the filtered denominator as
\begin{equation}
\label{eq:appendix_loss_sets}
\begin{aligned}
    \mathcal{A}(i) &= \{j\in\mathcal{B}\mid y_j=y_i,\ j\notin\mathcal{P}(i)\},\\
    \mathcal{D}(i) &= \mathcal{B}\setminus\mathcal{A}(i).
\end{aligned}
\end{equation}
With $s_{ij}=f_{r,i}^{\top}f_{t,j}/\tau$, define the contrastive term over a
candidate set $\mathcal{S}$ as
\begin{equation}
\label{eq:appendix_contrastive_term}
    \ell_i(p,\mathcal{S}) =
    -\log\frac{\exp(s_{ip})}
    {\sum_{j\in\mathcal{S}}\exp(s_{ij})}.
\end{equation}
The three per-query objectives are then
\begin{equation}
\label{eq:appendix_loss_comparison}
\begin{aligned}
    \mathcal{L}_{\mathrm{InfoNCE}}^{(i)}
    &= \ell_i(p_i,\mathcal{B}),\\
    \mathcal{L}_{\mathrm{SupCon}}^{(i)}
    &= \frac{1}{|\mathcal{P}(i)|}
       \sum_{p\in\mathcal{P}(i)}\ell_i(p,\mathcal{B}),\\
    \mathcal{L}_{\mathrm{CoGR}}^{(i)}
    &= \frac{1}{|\mathcal{P}(i)|}
       \sum_{p\in\mathcal{P}(i)}\ell_i(p,\mathcal{D}(i)).
\end{aligned}
\end{equation}
InfoNCE therefore uses one designated positive and treats every other target
as a negative. SupConLoss expands the positive pool to $\mathcal{P}(i)$ but
keeps the full batch in the denominator, so targets in $\mathcal{A}(i)$ remain
hard negatives. The CoGR loss retains the multi-positive supervision while
removing $\mathcal{A}(i)$ from the denominator. All objectives are averaged
over the queries in the batch. Table~\ref{tab:appendix_loss_ablation} reports
the resulting performance.

\begin{table}[t]
    \centering

    \footnotesize
    \setlength{\tabcolsep}{3.2pt}
    \begin{tabular}{@{}lcccc@{}}
        \toprule
        Objective & R@1 & SC-R$_a$@1 & SC-R$_v$@1 & SC-R$_c$@1 \\
        \midrule
        InfoNCE & 83.31 & 81.58 & 94.39 & 78.29 \\
        SupConLoss & 83.60 & 82.89 & 94.62 & 78.33 \\
        CoGR loss & \textbf{83.61} & \textbf{83.31} & \textbf{96.56} & \textbf{78.36} \\
        \bottomrule
    \end{tabular}
    \caption{Contrastive-loss ablation on Language-Augmented
    CASIA-B (\%). The CoGR loss mainly improves viewpoint-specific recall, while its gains in overall R@1 and composite recall are marginal.}
    \label{tab:appendix_loss_ablation}
\end{table}

SupConLoss improves over InfoNCE by 0.29 pp in R@1 and by 1.31,
0.23, and 0.04 pp in attribute-only, viewpoint-only, and composite SC-R@1,
respectively. This result supports using all in-batch targets that satisfy the
CoGR relevance criterion as positives instead of supervising only the
designated target. The CoGR loss further improves over SupConLoss by 0.42 pp on
attribute-only SC-R@1 and 1.94 pp on viewpoint-only SC-R@1, while the gains in
R@1 (0.01 pp) and composite SC-R@1 (0.03 pp) are marginal. Relative to
InfoNCE, the corresponding gains are 0.30, 1.73, 2.17, and 0.07 pp. The
improvement is therefore concentrated in change-specific recall, especially
viewpoint compliance, rather than in a large shift in overall strict retrieval.
This pattern is consistent with the objective design: treating a
same-identity but condition-mismatched sequence as a hard negative introduces
a gradient that conflicts with PIA's identity anchor, whereas excluding it
preserves identity supervision without labeling it as a valid condition-level
match.
\section{Qualitative Results and Failure Analysis}
\label{sec:appendix_qualitative}

We provide successful and failed retrievals for attribute, viewpoint, and
composite condition changes.  Each visualization includes the instruction,
query, ground truth, and top-5 results, allowing identity preservation and
condition consistency to be inspected jointly.  Across all figures, blue and
orange borders identify the query and ground truth, while green and red
borders follow the relevance labels used by the evaluation.

\subsection{Successful Retrievals}

\begin{figure*}[t]
  \centering
  \includegraphics[width=0.96\textwidth]{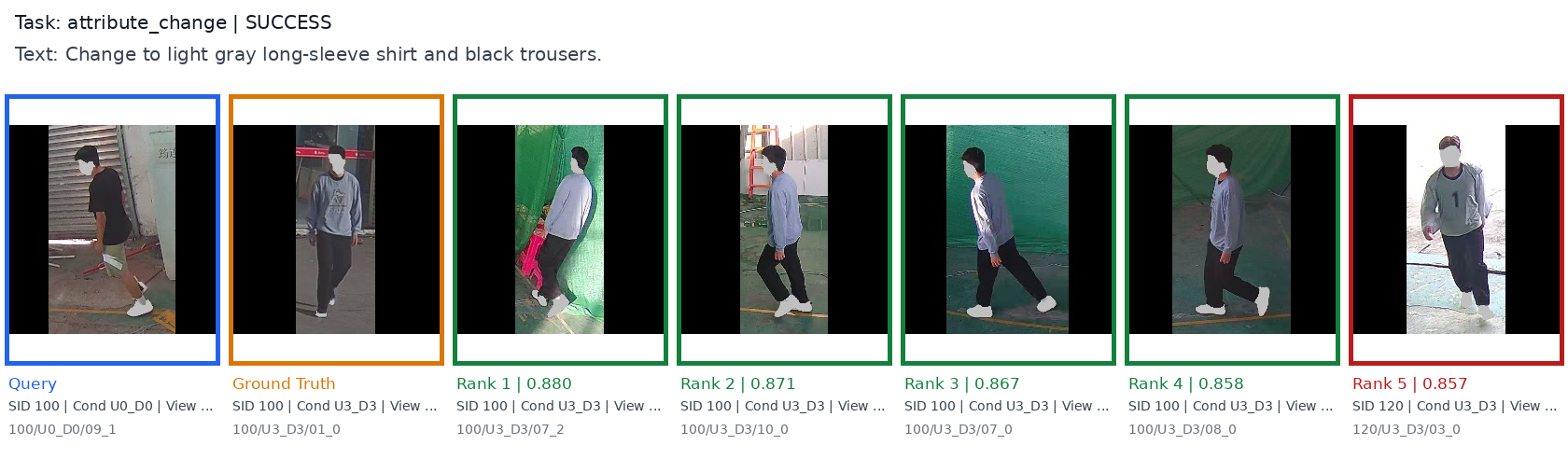}\\[0.1em]
  \includegraphics[width=0.96\textwidth]{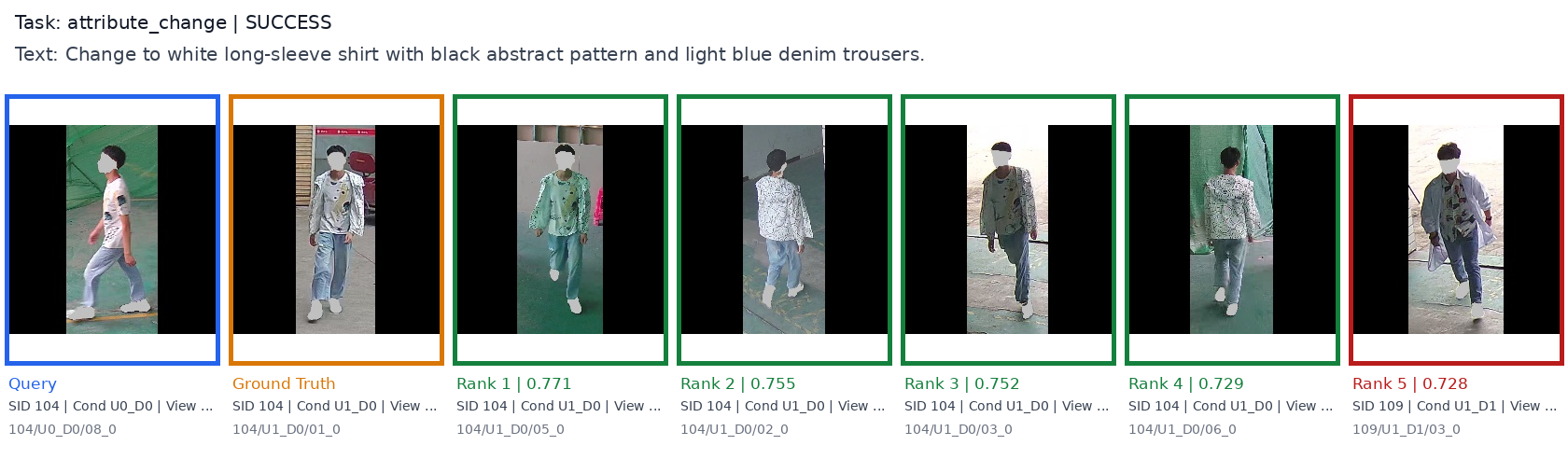}\\[0.1em]
  \includegraphics[width=0.96\textwidth]{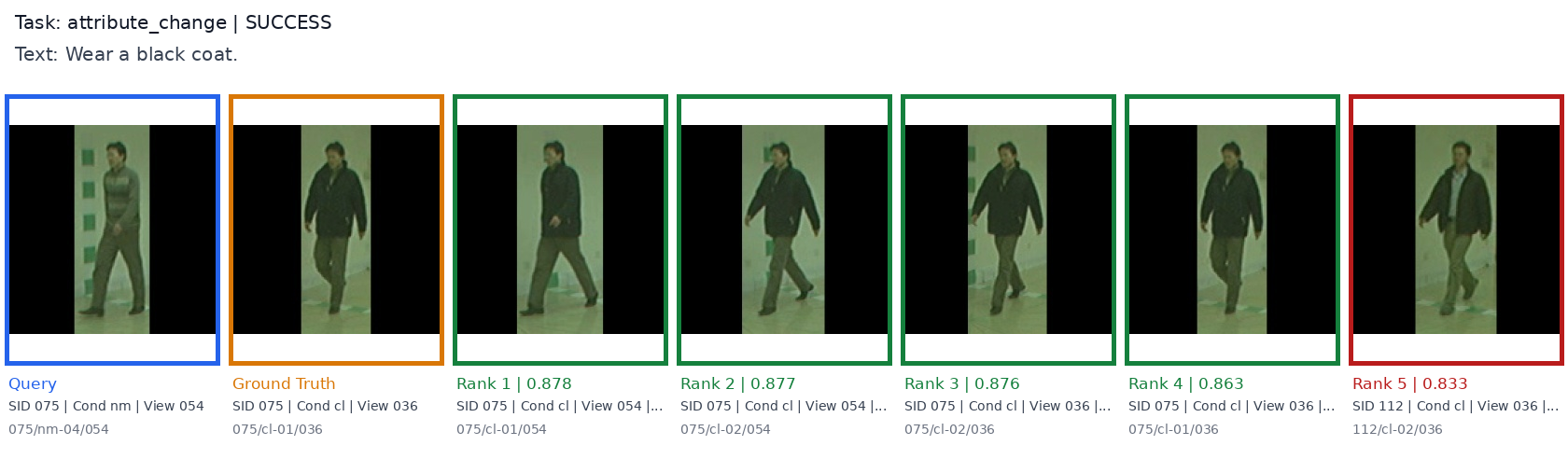}\\[0.1em]
  \includegraphics[width=0.96\textwidth]{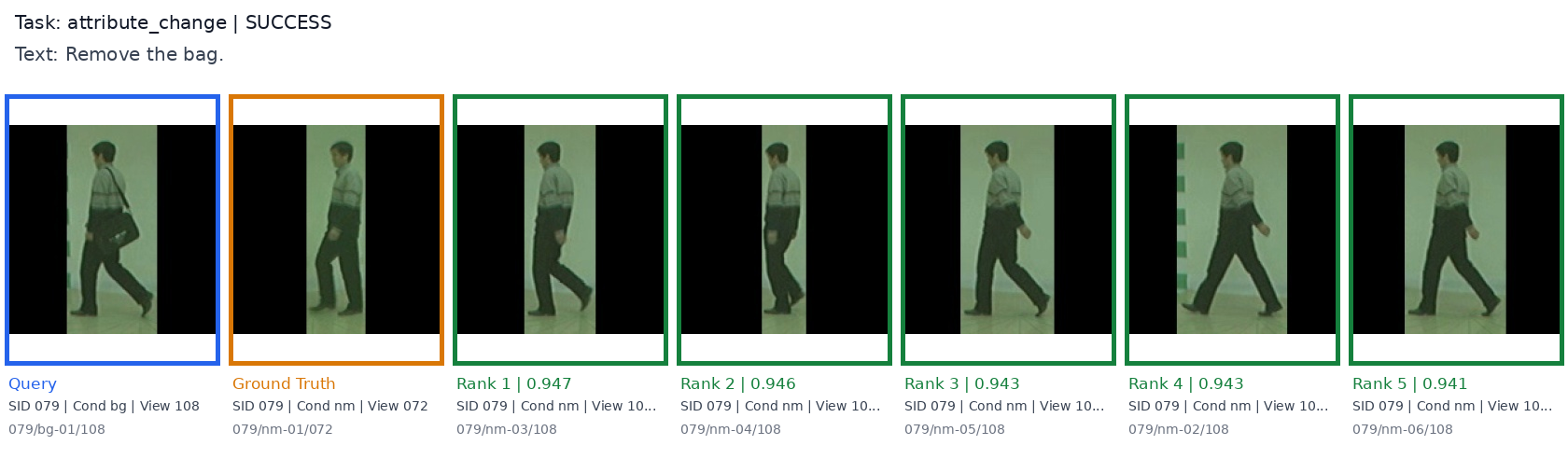}
  \caption{Successful attribute-conditioned retrievals on CCPG (top two
  rows) and CASIA-B (bottom two rows).  Each row contains the query (blue),
  ground truth (orange), and ranked candidates, with green and red borders
  following the evaluation labels.  The examples cover clothing color and
  pattern changes, adding a coat, and removing a carried bag; a relevant
  result is ranked first in every row.}
  \label{fig:success_attribute_cases}
\end{figure*}

\begin{figure*}[t]
  \centering
  \includegraphics[width=0.96\textwidth]{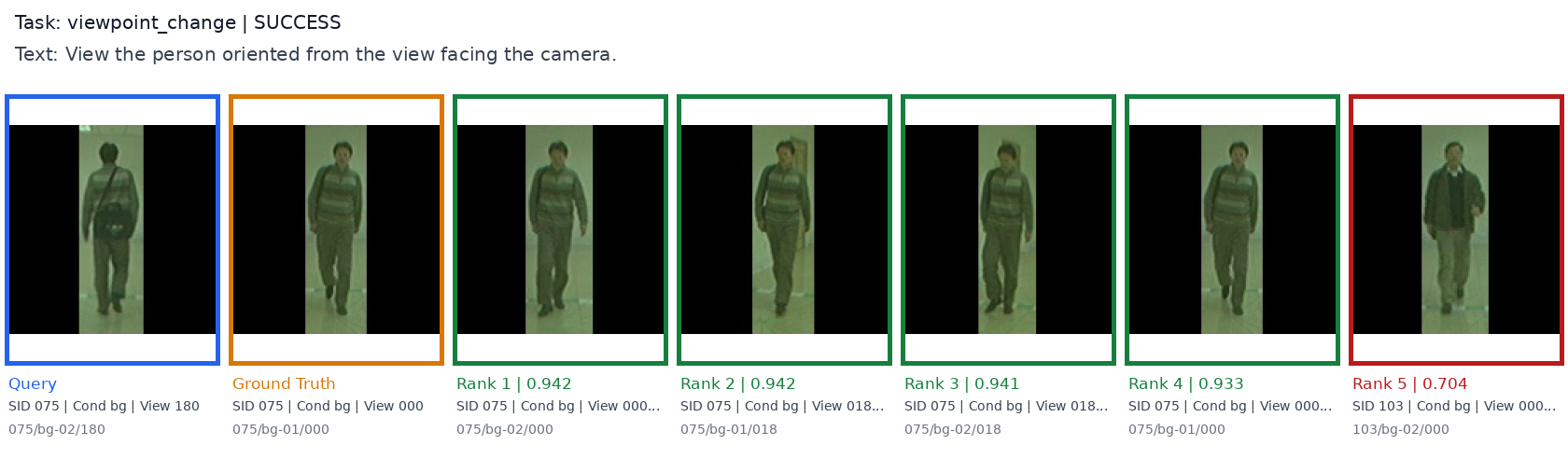}\\[0.1em]
  \includegraphics[width=0.96\textwidth]{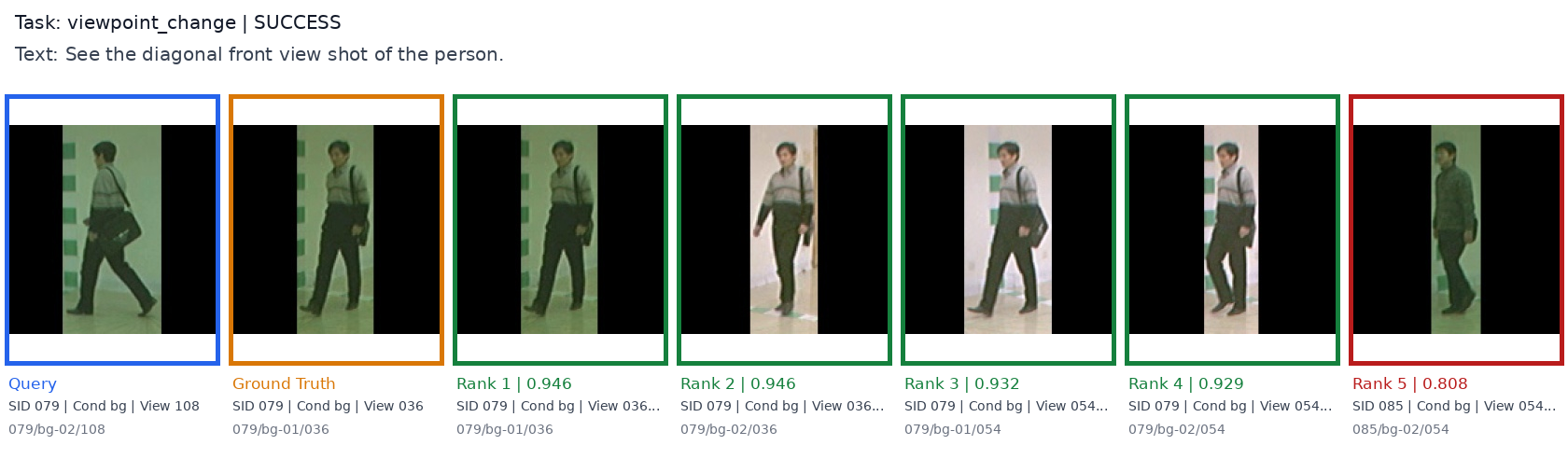}\\[0.1em]
  \includegraphics[width=0.96\textwidth]{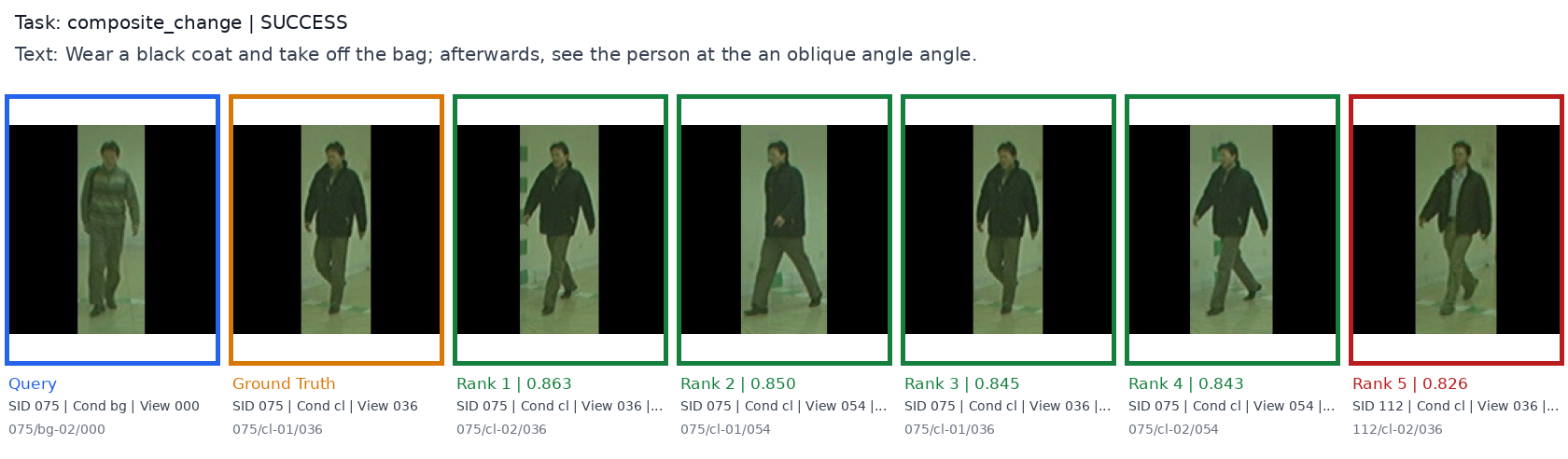}\\[0.1em]
  \includegraphics[width=0.96\textwidth]{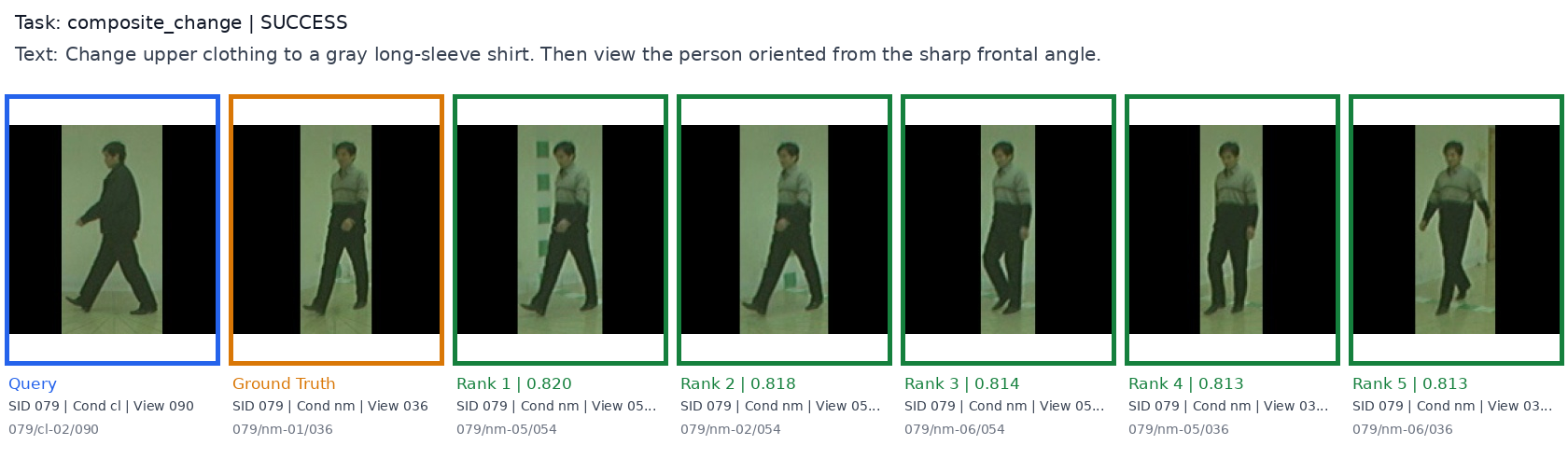}
  \caption{Successful viewpoint (top two rows) and composite condition
  retrievals (bottom two rows) on CASIA-B.  The viewpoint examples retrieve
  the requested frontal or front-oblique orientation while retaining
  identity.  The composite examples jointly change appearance, accessories,
  and viewpoint, and still place a relevant result at Rank~1.}
  \label{fig:success_view_composite_cases}
\end{figure*}

Figures~\ref{fig:success_attribute_cases} and
\ref{fig:success_view_composite_cases} show successful cases on CCPG and
CASIA-B.  We include two examples for each task family to cover both isolated
condition changes and multi-condition composition.

Across the successful cases, the top-ranked result follows the requested
condition without replacing the subject identity.  The attribute examples
span both large clothing changes and accessory removal, while the viewpoint
examples move between substantially different orientations.  The composite
examples further show that appearance and viewpoint constraints can be
satisfied together rather than being handled as independent single-condition
queries.

\subsection{Failure Cases}

\begin{figure*}[t]
  \centering
  \includegraphics[width=0.96\textwidth]{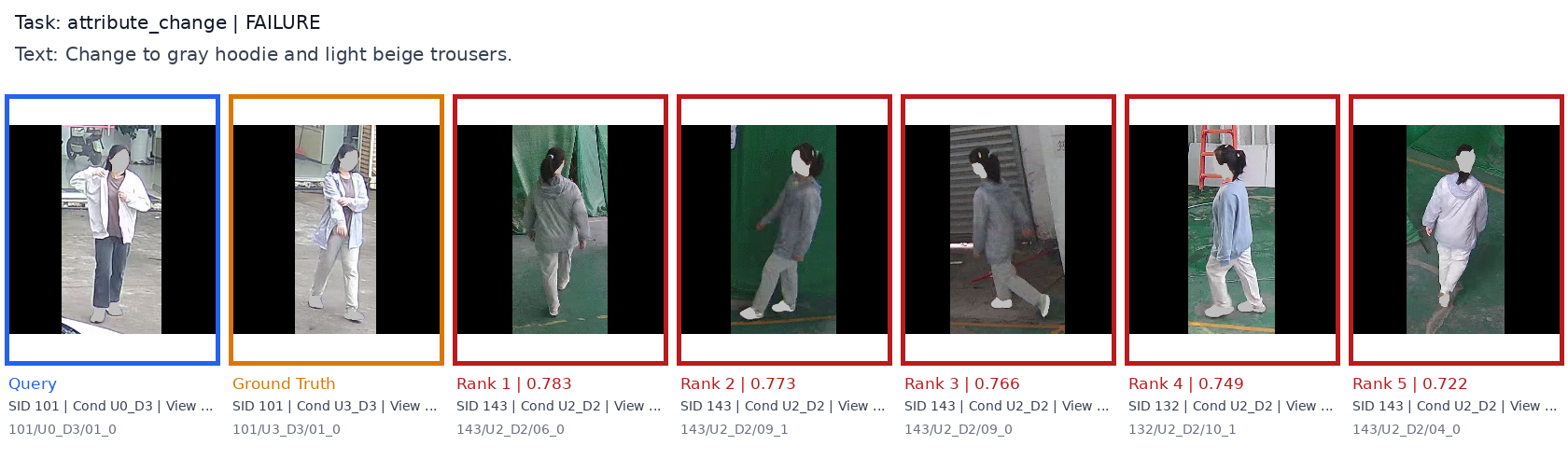}\\[0.1em]
  \includegraphics[width=0.96\textwidth]{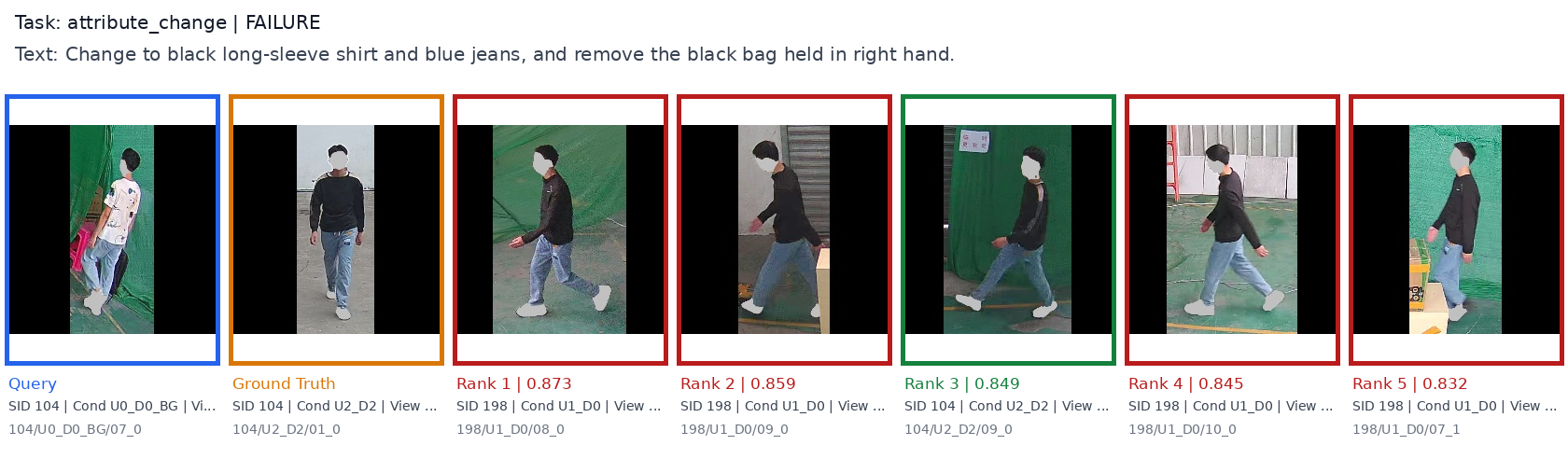}\\[0.1em]
  \includegraphics[width=0.96\textwidth]{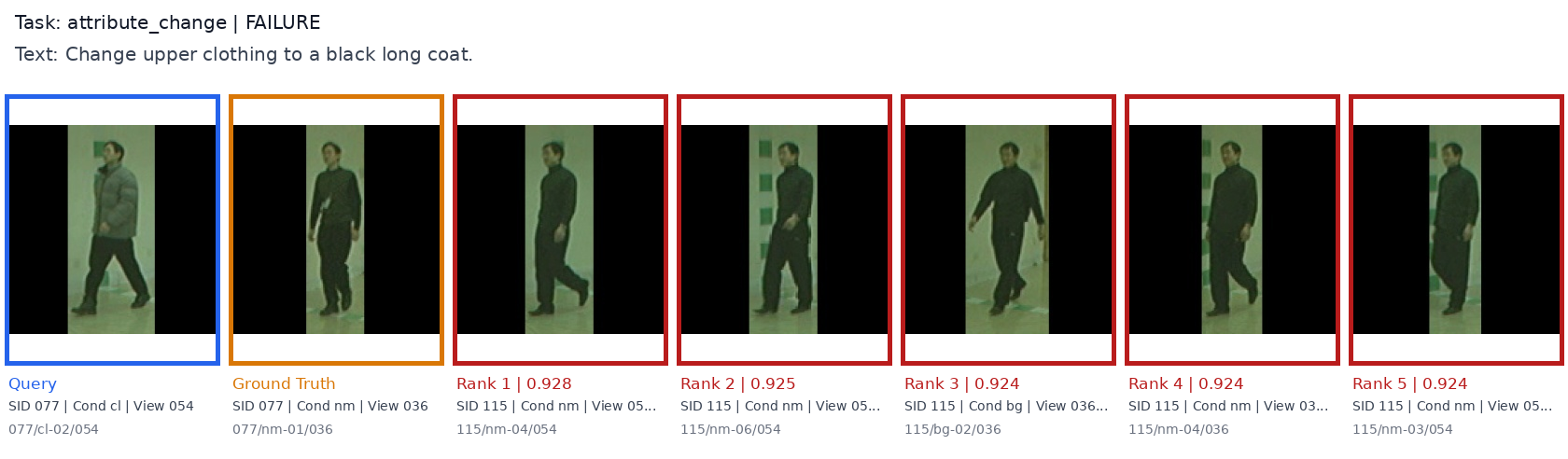}\\[0.1em]
  \includegraphics[width=0.96\textwidth]{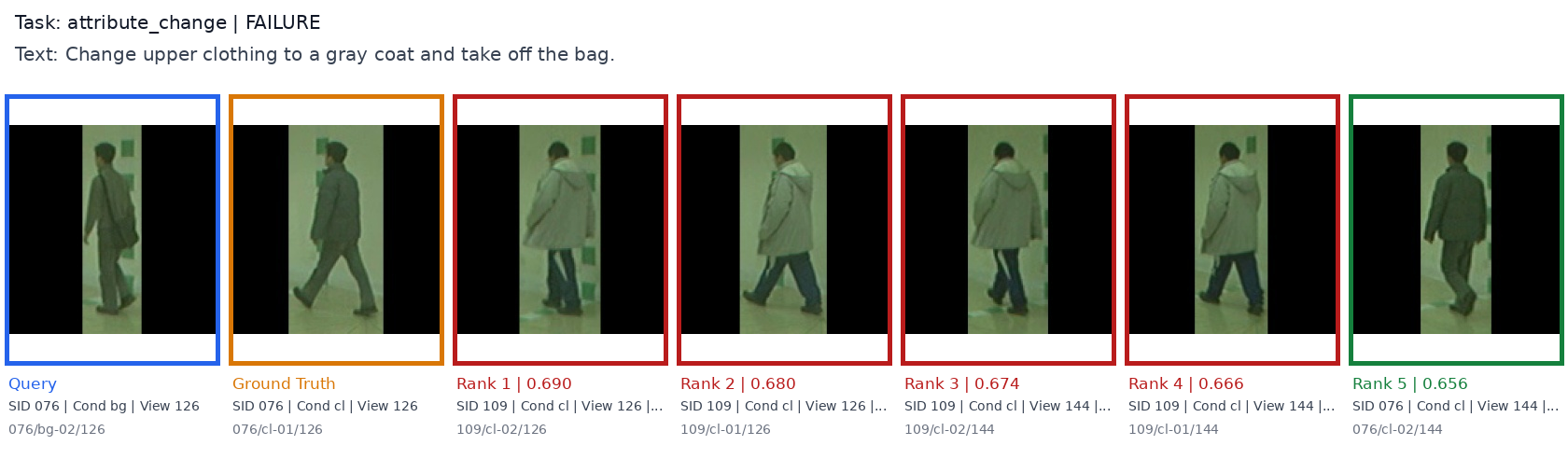}
  \caption{Attribute-change failures on CCPG (top two rows) and CASIA-B
  (bottom two rows).  In the first and third rows, all displayed candidates
  are invalid despite matching much of the requested appearance.  In the
  second and fourth rows, relevant sequences are present but are delayed until
  Rank~3 and Rank~5, respectively, behind candidates from other identities.}
  \label{fig:failure_attribute_cases}
\end{figure*}

\begin{figure*}[t]
  \centering
  \includegraphics[width=0.96\textwidth]{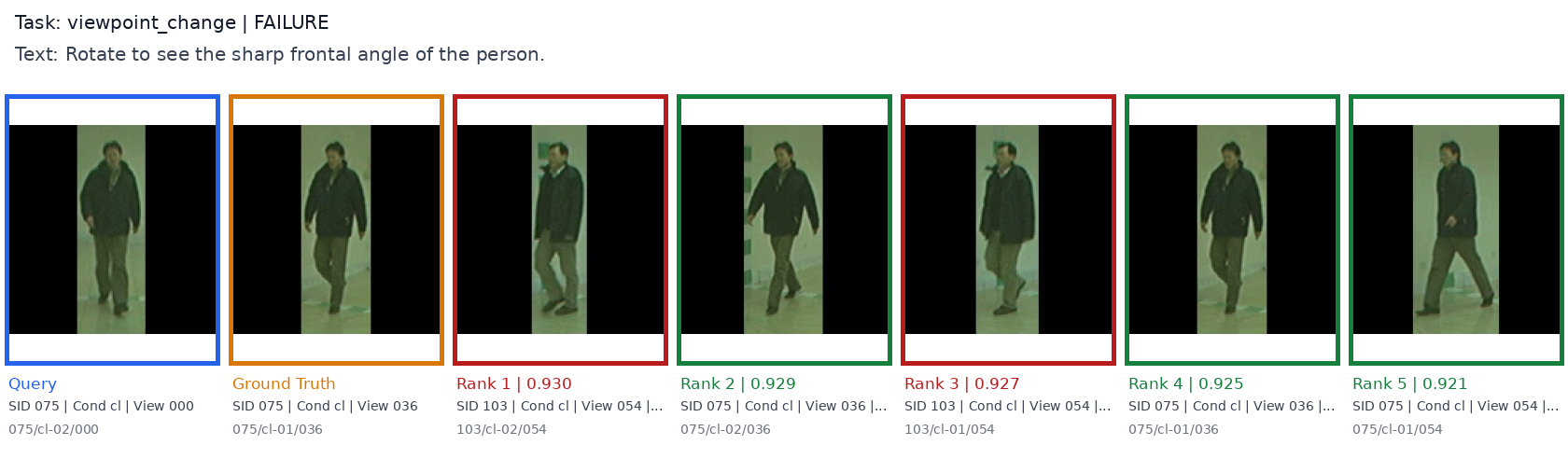}\\[0.1em]
  \includegraphics[width=0.96\textwidth]{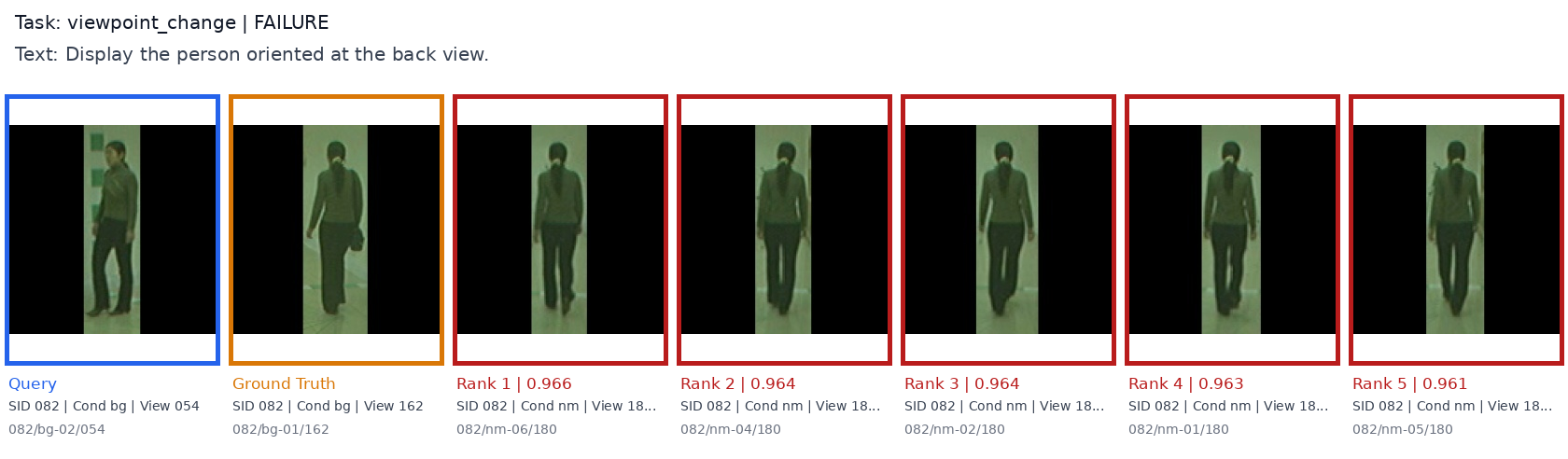}\\[0.1em]
  \includegraphics[width=0.96\textwidth]{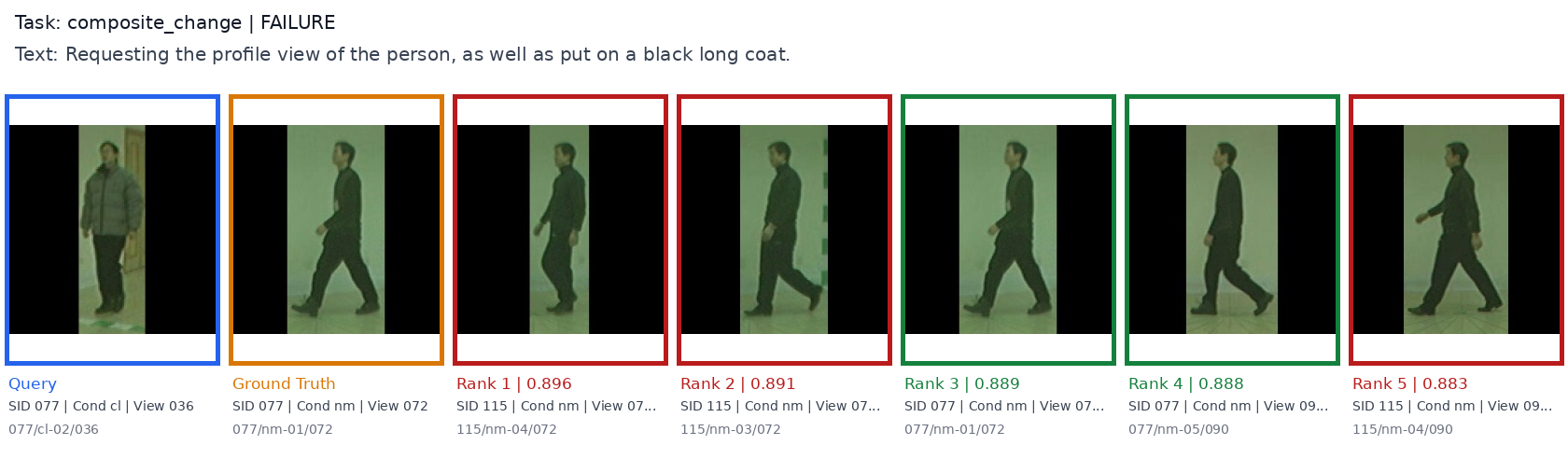}\\[0.1em]
  \includegraphics[width=0.96\textwidth]{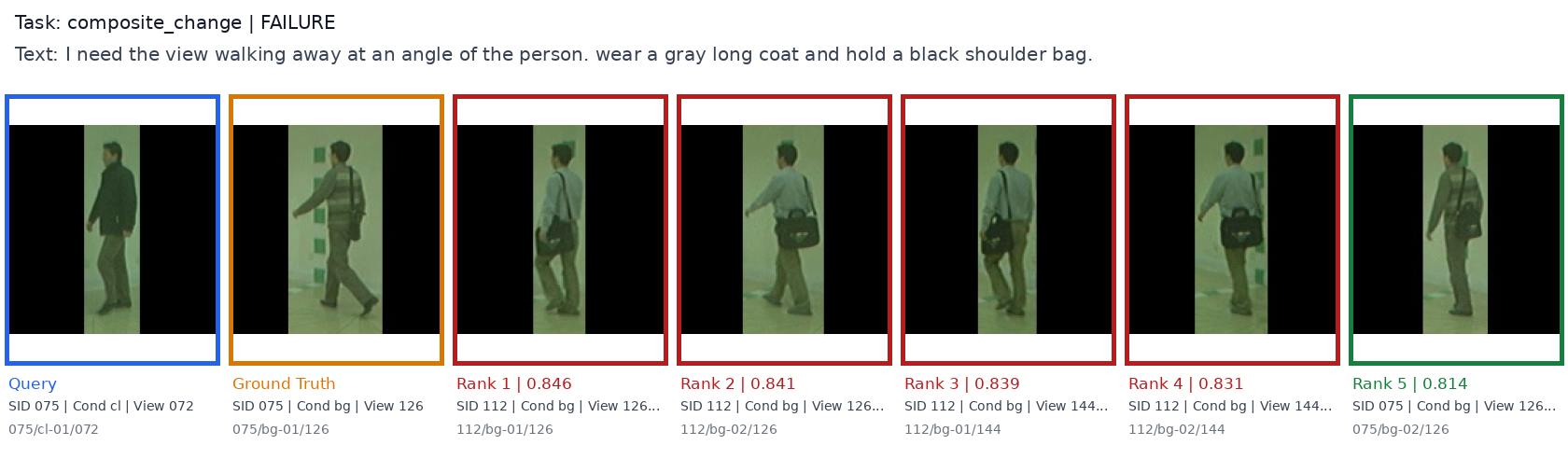}
  \caption{Failure cases for viewpoint-change instructions (top two rows) and
  composite-change instructions (bottom two rows) on CASIA-B.  In the first
  row, an invalid identity is ranked above a relevant result at Rank~2.  In the
  second, the retrieved sequences preserve identity and satisfy the requested
  back view, but fail to retain the original bag-carrying condition.  Under
  composite instructions, candidates satisfying the requested conditions but
  belonging to other identities outrank relevant results, which first appear
  at Rank~3 and Rank~5 in the bottom two rows.}
  \label{fig:failure_view_composite_cases}
\end{figure*}

\begin{figure*}[t]
  \centering
  \includegraphics[width=0.72\textwidth]{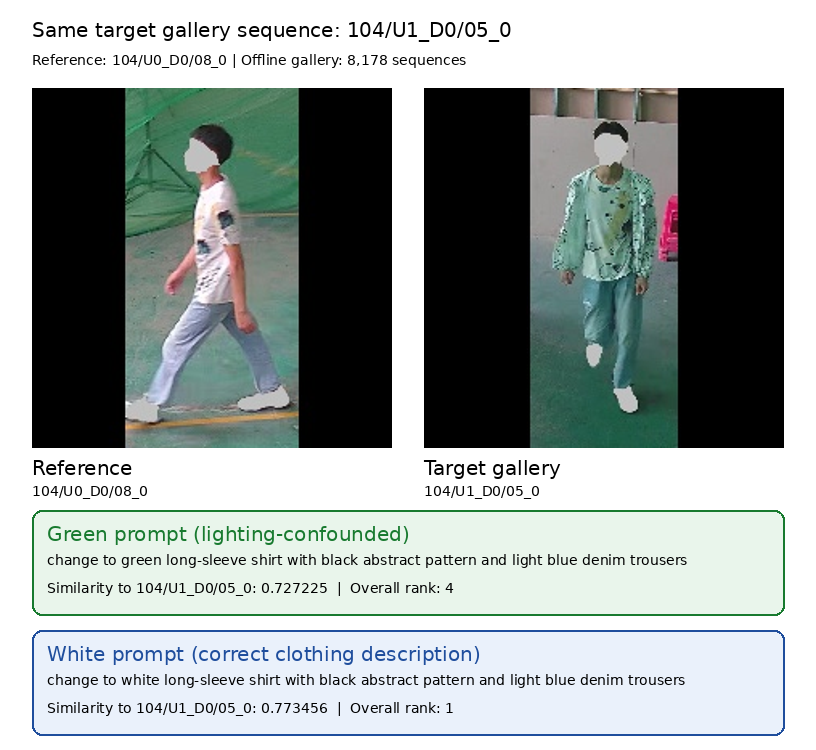}
  \caption{Lighting-confounded prompt comparison on CCPG.  The reference and
  target gallery sequence are fixed, and only the color word changes from the
  correct \emph{white} to \emph{green}.  The similarity decreases from
  0.77 to 0.73 and the target moves from Rank~1 to Rank~4.  This modest
  reduction is consistent with the green cast of the scene making the
  nominally white clothing appear greenish.}
  \label{fig:failure_lighting_confound}
\end{figure*}

Figures~\ref{fig:failure_attribute_cases} and
\ref{fig:failure_view_composite_cases} provide complementary failure cases.
They include both ranking errors, where a relevant sequence remains in the
top five, and retrieval errors for which no valid result appears in the shown
list. Figure~\ref{fig:failure_lighting_confound} further isolates prompt-color
sensitivity by fixing the reference and target sequence while changing only
the color word in the instruction.

The failures reveal three distinct limitations.  Attribute, viewpoint, and
composite instructions can produce identity drift when visually plausible
candidates from other subjects match the requested semantics more strongly.
Some valid targets remain in the top five, indicating that the embedding
contains useful identity evidence but the final ranking does not weight it
sufficiently.  The second viewpoint example exposes a different form of
condition leakage: the requested back view and subject identity are both
correct, but the bag-carrying condition changes from \texttt{bg} to
\texttt{nm}.  These observations motivate stronger identity-conditioned
scoring under large semantic changes and explicit preservation of conditions
that are not modified by the instruction.

Figure~\ref{fig:failure_lighting_confound} exposes an additional
environmental confound.  Replacing the correct color term \emph{white} with
the incorrect term \emph{green} reduces the target similarity by only
0.04.  Because clothing color is the modified condition, stricter
attribute grounding should penalize this mismatch more strongly.  However,
the green-dominated illumination and background tint the observed appearance
of the subject, making the incorrect prompt partially compatible with the
pixels.  This case suggests that the current representation may not fully
disentangle intrinsic clothing color from illumination-induced appearance. Future
work should evaluate color-conditioned retrieval under controlled changes in
illumination and color temperature, and investigate color-constancy or
illumination-invariant representations.

\section*{Ethical Statement}

This study uses the publicly available CCPG and CASIA-B
datasets and does not collect new human-subject data or
introduce additional identity annotations. We follow the
licenses and intended research use of the original datasets.
The language-augmented datasets contain only automatically
generated textual annotations, retrieval triplets, and data
splits derived from the original datasets.

Nevertheless, composed gait retrieval is a biometric
technology that could potentially be misused for
non-consensual identification, persistent tracking, or
large-scale surveillance. Its performance may also vary
across demographic groups, clothing conditions, physical
abilities, and capture environments that are unevenly
represented in the source datasets. We therefore recommend
that the generated annotations, retrieval splits, code, and
models be used only for legitimate research purposes and
subject to the licenses and access policies of the original
datasets. Any real-world deployment should require
appropriate authorization, privacy safeguards, human
oversight, and careful evaluation of demographic and
operational biases.

\end{document}